\documentclass{article}

 \usepackage[preprint]{neurips_2026}

\usepackage[utf8]{inputenc} % allow utf-8 input
\usepackage[T1]{fontenc}    % use 8-bit T1 fonts
\usepackage{hyperref}       % hyperlinks
\usepackage{url}            % simple URL typesetting
\usepackage{booktabs}       % professional-quality tables
\usepackage{amsfonts}       % blackboard math symbols
\usepackage{nicefrac}       % compact symbols for 1/2, etc.
\usepackage{microtype}      % microtypography

\usepackage{xcolor}         % colors

\usepackage{amsmath} 
\usepackage{wrapfig}
\usepackage{graphicx}
\usepackage{enumitem}
\usepackage{bm}
\usepackage{mathtools}
\usepackage{subcaption}
\usepackage{booktabs}
\usepackage{xcolor}
\usepackage{multirow}
\usepackage[table]{xcolor}
\usepackage[most]{tcolorbox}
\usepackage{algorithm}
\usepackage{algpseudocode}
\definecolor{darkgreen}{rgb}{0.0, 0.5, 0.0}

\definecolor{mainbg}{RGB}{250, 250, 250}
\definecolor{headergray}{RGB}{80, 80, 80}
\definecolor{correctGreen}{RGB}{0, 128, 0}
\definecolor{grpoBlue}{RGB}{0, 0, 200}
\title{PRPO: Perception-Reinforced Policy Optimization via Token-Level Dynamic Advantage Reshaping}

\author{
    Qiming Li$^{1,\spadesuit}$, Tianlun Li$^{1,\dagger}$, Xiaolong Cheng$^{1}$,\\
    \textbf{Hangyu Li$^{2}$, Ruiyan Gong$^{1}$, Kangning Niu$^{1}$,  Kaitao Jiang$^{1}$, Mu Xu$^{1}$}\\
    $^{1}$Amap CV Lab, Alibaba Group,\ \ \  $^{2}$Peking University \\
    \texttt{qimingli1104@gmail.com}
}

\begin{document}

\maketitle

% === 插入作者标志的注脚 ===
{
\let\thefootnote\relax
\footnotetext{$^\spadesuit$ Work done during an internship at Amap CV Lab, Alibaba Group.}
\footnotetext{$^\dagger$ \ Corresponding author.}
}
% ==========================

\begin{abstract}
Reinforcement Learning with Verifiable Rewards (RLVR) has become an effective paradigm for improving the reasoning capability of Large Vision-Language Models (LVLMs). However, existing RLVR methods primarily rely on trajectory-level outcome rewards, which assign identical learning signals across all generated tokens. This coarse-grained credit assignment is fundamentally mismatched to multimodal reasoning, where only a sparse subset of tokens is causally grounded in visual evidence. Consequently, these pivotal perceptual tokens receive weak supervision and are often overwhelmed by language priors or reasoning-template tokens. To address this limitation, we propose \textbf{Perception-Reinforced Policy Optimization (PRPO)}, a token-level reinforcement learning framework that explicitly identifies and reinforces pivotal perceptual tokens within long-horizon multimodal reasoning trajectories. PRPO introduces \textbf{Robust Visual Dependency (RVD)}, a principled metric that identifies tokens whose predictions are both visually grounded and perturbation-stable, filtering out brittle or noisy visual tokens. Based on RVD, we further propose \textbf{Perceptual Advantage Reshaping (PAR)}, a token-level credit assignment technique that amplifies perceptually informative tokens while preserving stable gradients for non-perceptual tokens. Extensive experiments on seven multimodal reasoning benchmarks demonstrate that \textbf{PRPO} consistently outperforms strong LVLM baselines across both 3B and 7B model scales, achieving average gains of 23.3\% and 21.1\%, respectively. \textbf{PRPO} achieves state-of-the-art performance with improved training efficiency and stronger cross-task generalization. Our findings highlight the importance of fine-grained credit assignment for scalable multimodal reinforcement learning.
\end{abstract}

\section{Introduction}
Recent advances in reinforcement learning from verifiable rewards (RLVR) have significantly strengthened the reasoning capability of Large Language Models (LLMs), particularly when paired with online policy optimization algorithms such as Group Relative Policy Optimization (GRPO) \citep{shao2024deepseekmath, guo2025deepseek, openai2024learningtoreason, team2025kimi, yang2025qwen3, anthropic2025claude, zhang2025survey, qu2025survey, yan2025learning}. These gains have been most evident in text-focused tasks and benchmarks. 
Inspired by this success, researchers have increasingly explored how to adapt RLVR to Large Vision-Language Models (LVLMs). 
Current efforts can be broadly organized into three lines of work: methods that improve training data quality or composition~\citep{li2025truth, liang2025modomodo, liu2025noisyrollout, yao2025r1, chen2025g1, meng2025mm, huang2025vision, yang2025r1}; methods centered on reward modeling, reward shaping, or reward-system design~\citep{shen2025vlm, xia2025visionary, wang2025skywork, xiao2025advancing, yu2025perception, wan2025srpo, he2025framethinker}; and methods that modify the underlying optimization procedure itself~\citep{wang2025vl, zhao2025absolute}.

\begin{figure*}
\centering
  \includegraphics[width=1.0\linewidth]{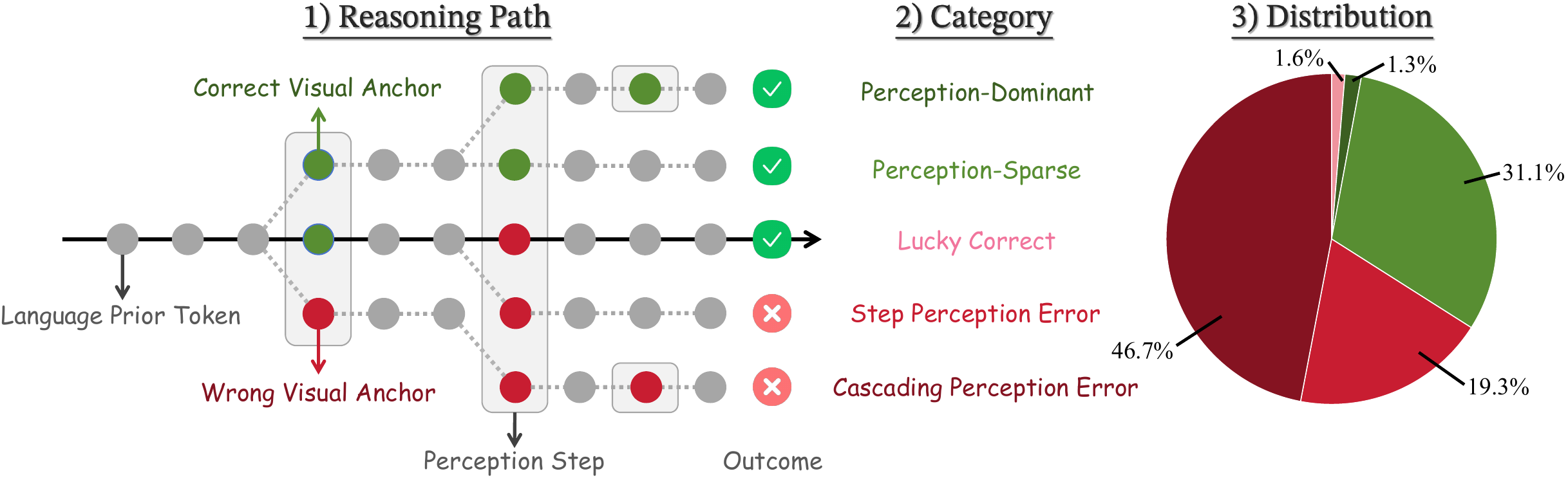}
  \caption{Error analysis of multimodal reasoning failures using Qwen2.5-VL-7B on the MathVerse benchmark. Analysis results indicate that LVLMs are limited by perceptual bottlenecks within reasoning trajectories rather than insufficient reasoning capabilities.}
  \label{error}
\end{figure*}

Despite these advances, existing multimodal RLVR methods largely inherit designs from language-only settings, which assign a single trajectory-level reward to the final response and broadcast the same optimization signal across all generated tokens. While this design is useful for purely textual reasoning, it becomes fundamentally mismatched in multimodal reasoning. To better characterize this mismatch, we conduct an error analysis on MathVerse~\citep{zhang2024mathverse} using Qwen2.5-VL-7B. As shown in Figure~\ref{error}, the majority of failures do not stem from insufficient reasoning capability, but from errors in visual perception occurring within long reasoning trajectories. Specifically, LVLMs often ground on incorrect visual regions, misread local evidence, or extract erroneous numerical cues at intermediate perception steps. Such behaviors either lead to isolated \emph{Step Perception Errors} or accumulate into  \emph{Cascading Perception Errors}. In contrast, failures attributable purely to reasoning errors are relatively rare. Furthermore, we find that key perceptual tokens in multimodal reasoning (e.g., coordinates, numerical values, or geometric cues) are both sparse and highly influential, yet they receive the same supervision as non-perceptual linguistic tokens under standard RL objectives.
We identify this mismatch as the \textbf{\textit{Perceptual Credit Assignment Problem}}, which contains two key issues: (i) sparse perceptual signals cannot be explicitly identified or reinforced from trajectory-level rewards, and (ii) uniform advantage assignment causes gradient dilution, where critical perceptual signals are overwhelmed by abundant reasoning tokens. As a result, existing RL methods tend to optimize surface-level reasoning fluency rather than improving visually grounded decision-making.

To address this problem, we propose \textbf{Perception-Reinforced Policy Optimization (PRPO)}, a dynamic token-level reinforcement learning framework that explicitly targets perceptual credit assignment in multimodal reasoning. \textbf{PRPO} contains two key components: \textbf{(i) Robust Visual Dependency (RVD)}, identifies tokens that are genuinely grounded in visual evidence and robust to perturbations, and \textbf{(ii) Perceptual Advantage Reshaping (PAR)}, which reshapes trajectory-level advantage into fine-grained token-level learning signals based on RVD, amplifying informative perceptual tokens while preserving stable optimization gradients on reasoning-template tokens.

Extensive experiments on seven multimodal reasoning benchmarks demonstrate that \textbf{PRPO} consistently improves performance across both 3B and 7B models, with average gains of 23.3\% and 21.1\%, respectively. Moreover, \textbf{PRPO} exhibits stronger performance under the same training budget and cross-task generalization, indicating that fine-grained perceptual reinforcement is more effective than uniform trajectory-level optimization in multimodal RLVR context.

Our contributions are three-fold:
\begin{itemize}[leftmargin=*]
    \item We identify \textbf{\emph{Perceptual Credit Assignment}} as a key bottleneck in multimodal RLVR: most failures arise from sparse but critical perception errors rather than insufficient reasoning capability, while existing trajectory-level rewards fail to provide effective supervision to pivotal perceptual tokens.
    \item We propose \textbf{PRPO}, a token-level RL framework with \textbf{RVD} and \textbf{PAR} technique to identify and reinforce pivotal perceptual tokens while preserving stable optimization gradients.
    \item We achieve state-of-the-art (SOTA) results across seven multimodal reasoning benchmarks, demonstrating the importance of fine-grained credit assignment for multimodal reinforcement learning.
\end{itemize}

\section{Related Work}

% \paragraph{Multimodal Reasoning.}
% Although Large Language Models (LLMs) have demonstrated exceptional reasoning capabilities within pure-text contexts~\citep{guo2025deepseek}, extending this proficiency to Large Vision-Language Models (LVLMs)~\citep{bai2025qwen2, hurst2024gpt, team2024gemini, bai2025intern} remains challenging. Current multimodal architectures still struggle to match the reasoning depth seen in their unimodal counterparts~\citep{wang2024enhancing, dong2025insight,su2025openthinkimg,su2025thinking,shen2025satori,qu2024alleviating}. Overcoming this discrepancy demands the adaptation of foundational reinforcement learning techniques—such as PPO~\citep{schulman2017proximal} and GRPO~\citep{shao2024deepseekmath}—to accommodate the distinct characteristics of multimodal inputs.

\paragraph{Dominant Strategies in Multimodal RL.}
Current efforts to enhance multimodal RL largely focus on augmenting elements outside the core optimization algorithm. These include data-driven methods emphasizing the construction of visually-grounded datasets~\citep{bai2025univg, li2025truth, liang2025modomodo} and CoT distillation~\citep{chen2025vinci, huang2025vision, meng2025mm}, as well as reward-centric methodologies designing improved feedback mechanisms~\citep{ma2025one, fan2025grit, xia2025visionary}. Recognizing the vision-language gap, recent works like Perception-Aware Policy Optimization (PAPO)~\citep{wang2025perception} explicitly account for modality differences to engineer perception-sensitive signals. However, even with such modality-aware designs or modality-agnostic algorithmic advances like DAPO~\citep{yu2025dapo}, these strategies still broadcast a uniform learning signal across the entire sequence. This indiscriminate credit assignment creates a severe bottleneck where critical, sparse visual signals are easily submerged by high-volume linguistic priors.

\paragraph{Pivotal Tokens Identification in LVLMs.}
To mitigate gradient dilution, it is crucial to isolate specific perceptual tokens from critical reasoning. In text-only domains, this involves identifying high-entropy ``forking points''~\citep{wang2025beyond} or low-confidence errors~\citep{vassoyan2025ignore}. For LVLMs, approaches like VPPO~\citep{huang2025spotlight} attempt to address this by utilizing KL Divergence-based visual dependency to reweight trajectory-level advantages, aiming to focus gradient updates on visually-grounded steps. However, relying on a single visual dependency metric risks amplifying noise-sensitive tokens, and trajectory-level reward reshaping is fundamentally insufficient to precisely reinforce visually important tokens. To tackle this, our PRPO framework introduces a precise dynamic token-level optimization paradigm. By jointly evaluating visual dependency and consistency, we systematically filter out stochastic noise to isolate genuinely pivotal perceptual tokens. We then dynamically reshape their learning advantages, explicitly amplifying critical perceptual signals without triggering gradient collapse or compromising foundational reasoning capabilities.

\section{Preliminaries: Group-based Policy Optimization}
\label{sec:preliminary}

In multimodal reinforcement learning from verifier feedback (RLVF), we consider a dataset $\mathcal{D}$ where each sample consists of a visual input $I$, a textual query $q$, and a ground-truth answer $a$. Let $\pi_{\theta}$ denote the policy parameters of a LVLM.

\paragraph{Group Relative Policy Optimization (GRPO)}
GRPO~\citep{shao2024deepseekmath} serves as a value-model-free baseline that estimates advantages through intra-group relative performance. For a prompt $(I, q)$, the model generates a group of $G$ responses $\{o_i\}_{i=1}^{G}$, each receiving a binary outcome reward $R_i \in \{0, 1\}$ based on its equivalence to $a$. The advantage $\hat{A}_i$ is computed by normalizing rewards within the group:
\begin{equation}
\hat{A}_i = \frac{R_i - \text{mean}(\{R_k\}_{k=1}^{G})}{\text{std}(\{R_k\}_{k=1}^{G}) + \eta}, \quad \text{where } R_i = 
    \begin{cases} 
      1.0, & \text{if } \text{eq}(a, o_i) \\
      0.0, & \text{otherwise}
    \end{cases}
\label{eq: grpo advantage}
\end{equation}
where $\eta$ is a small constant for numerical stability. The objective is to maximize the clipped surrogate loss with an integrated KL penalty:
\begin{align}
\mathcal{J}_{\text{GRPO}}(\theta) = \mathbb{E} \left[ \frac{1}{G} \sum_{i=1}^{G} \left( \frac{1}{|o_i|} \sum_{t=1}^{|o_i|} \min \left( r^i_t(\theta) \hat{A}_i, \text{clip}(r^i_t(\theta), 1-\epsilon, 1+\epsilon) \hat{A}_i \right)  \right) \right] \nonumber
\end{align}
%  - \beta D_{\text{KL}}[\pi_{\theta} \| \pi_{\text{ref}}]
\begin{align}
    \text{with} \ \ r^i_t(\theta) = \frac{\pi_{\theta}(o^i_t | I, q, o_{<t})}{\pi_{\theta_{\text{old}}}(o^i_t | I,q, o_{<t})}.
\end{align}

\paragraph{Decoupled Clip and Dynamic Sampling (DAPO)}
DAPO~\citep{yu2025dapo} refines the optimization landscape by introducing an asymmetric clipping mechanism and dynamic sampling strategies. It utilizes a token-level objective with decoupled thresholds $(\epsilon_{\text{low}}, \epsilon_{\text{high}})$ and global length normalization:
\begin{align}
\mathcal{J}_{\text{DAPO}}(\theta) = & \mathbb{E} \Bigg[
\frac{1}{\sum_{i=1}^{G} |o_i|} 
\sum_{i=1}^{G} \sum_{t=1}^{|o_i|} 
\min \Big( 
r_{t}^i(\theta) \hat{A}_i,\, \text{clip}\big(r_{t}^i(\theta),
1 - \epsilon_{\text{low}},  1 + \epsilon_{\text{high}} \big) \hat{A}_i 
\Big) 
\Bigg]
\label{eq: dapo objective}
\end{align}
% - \beta D_{\text{KL}}[\pi_{\theta} \| \pi_{\text{ref}}]
\section{PRPO: Perception-Reinforced Policy Optimization}
\label{sec:method}

\begin{figure*}[!ht]
\centering
  \includegraphics[width=1.0\linewidth]{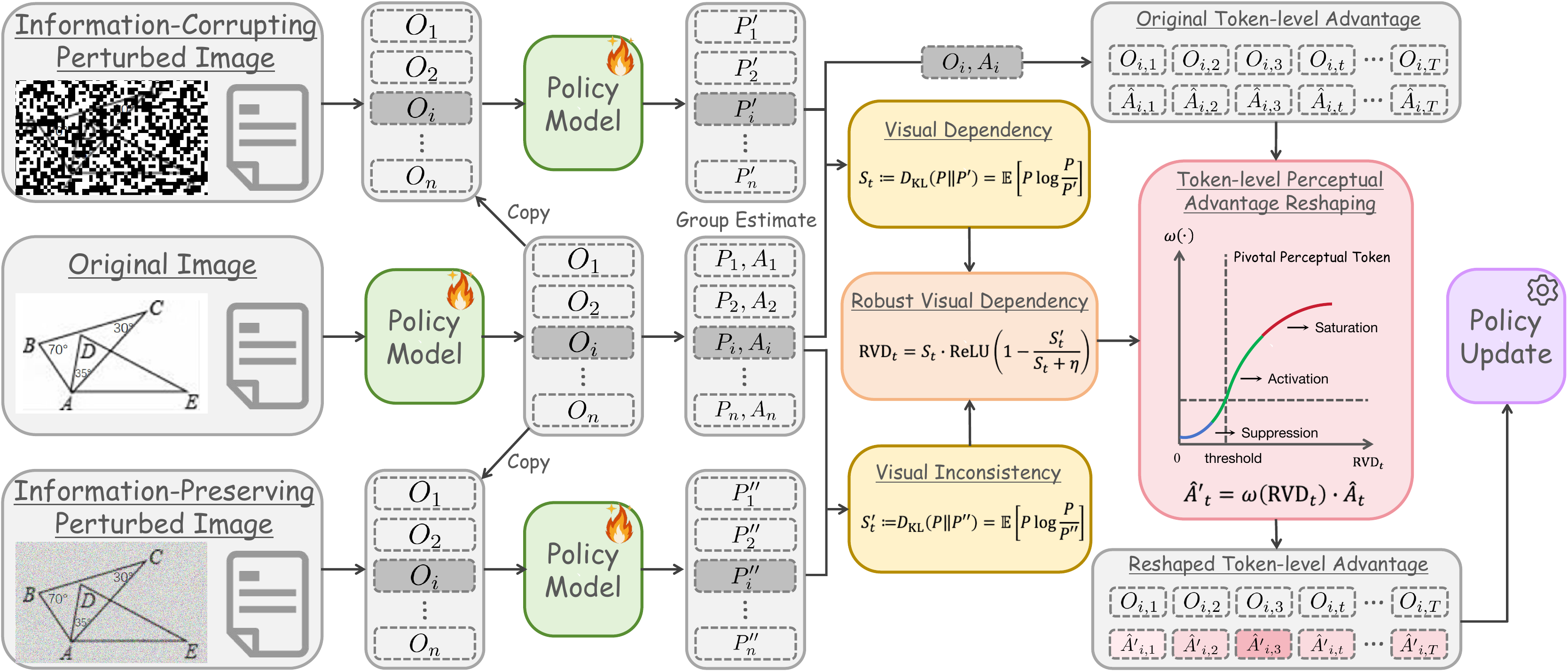}
  \caption{Overview of our proposed PRPO framework. PRPO consists of two core components: \textbf{(i) Robust Visual Dependency (RVD)}, which identifies \textit{pivotal perceptual tokens} through fine-grained trajectory analysis, and \textbf{(ii) Perceptual Advantage Reshaping (PAR)} algorithm, which applies a synaptic S-shaped curve to reshape uniform trajectory-level advantages, dynamically amplifying pivotal perceptual tokens while preserving stable optimization gradients for reasoning tokens.
  % PRPO consists of two key components: \textbf{(i) Robust Visual Dependency (RVD)}, which evaluates Visual Dependency ($S_t$) and Inconsistency ($S'_t$) across distinct dimensions via image perturbations, explicitly identifying sparse pivotal perceptual tokens and filtering out brittle visual noise; \textbf{(ii) Perceptual Advantage Reshaping (PAR)}, which applies a synaptic S-shaped curve to reshape uniform trajectory-level advantages, dynamically amplifying pivotal perceptual tokens while preserving stable optimization gradients for reasoning tokens.
  }
\end{figure*}

% \subsection{Motivation: The Perceptual Credit Assignment Problem}
% A fundamental limitation of existing outcome-based rewards (e.g., GRPO and DAPO) is the \textit{uniform broadcasting} of the sequence-level advantage $\hat{A}_i$ to all tokens. In multimodal reasoning, this creates a critical \textbf{Perceptual Credit Assignment} bottleneck. This manifests in two critical dimensions:
% \begin{itemize}[leftmargin=*]
%     \item \textbf{Sparsity of Perceptual Reward Signals:} In long-horizon reasoning, \textit{pivotal perceptual tokens} are exceptionally sparse. A trajectory-level reward inherently lacks the information capacity to localize and reinforce these rare perceptual anchors, treating them from rote linguistic transitions.
%     \item \textbf{Token-Level Gradient Dilution:} The advantage $\hat{A}_{i,t}$ is applied indiscriminately to each token in the trajectories. This "broadcasting" mechanism fails to isolate and reinforce the sparse set of \textit{pivotal perceptual tokens}—such as specific coordinates, numerical values, or geometric properties—that are truly essential for visually-grounded reasoning. Consequently, the critical perceptual signal is often overwhelmed by the high volume of abundant reasoning tokens.
% \end{itemize}

% In this work, we propose \textbf{Perception-Reinforced Policy Optimization (PRPO)}, a framework designed to identify and reinforce \textit{pivotal perceptual tokens} within long multimodal reasoning trajectories. 

\subsection{Robust Visual Dependency (RVD)}

In this section, we detail the methodology for identifying \textit{pivotal perceptual tokens} within long multimodal reasoning trajectories. Following established paradigms~\citep{huang2025spotlight}, we adopt Kullback-Leibler (KL) divergence as a principled heuristic metric to quantify the sensitivity of individual tokens to visual evidence, serving as the foundation for our robust dependency analysis.

\paragraph{Definition 4.1 (Visual Dependency).} 

\textit{Let $\mathcal{P}_{\text{str}}$ be a strong perturbation operator that maps $I$ to a state $\tilde{I}_{\text{str}}$ where the visual information is severely corrupted (e.g., via a stochastic black mask). We define the Visual Dependency $S_t$ to formally quantify the policy's dependency on visual information for a specific output token $o_t$:}

\begin{equation}
    S_t \coloneqq D_{\text{KL}} \big( \pi_{\theta}(\cdot \mid I, q, o_{<t}) \, \| \, \pi_{\theta}(\cdot \mid \mathcal{P}_{\text{str}}(I), q, o_{<t}) \big)
\end{equation}

$S_t$ measures the absolute magnitude of the visual signal required for prediction. A high $S_t$ indicates that the visual input $I$ serves as a necessary perception anchor for the generation of token $o_t$, distinguishing visual grounded tokens  from pure language priors.

\paragraph{Definition 4.2 (Visual Inconsistency).} \textit{Let $\mathcal{P}_{\text{wk}}$ be a weak perturbation operator that introduces an infinitesimal stochastic jitter $\bm{\delta} \sim \mathcal{N}(0, \sigma^2 \cdot {I})$ to the pixel space. Crucially, $\mathcal{P}_{\text{wk}}$ ensures that the visual semantics remain strictly invariant.
We define the Visual Inconsistency $S'_t$ to measure the model's sensitivity to numerical noise for a specific output token $o_t$:}
\begin{equation}
    S'_t \coloneqq D_{\text{KL}} \big( \pi_{\theta}(\cdot \mid I, q, o_{<t}) \, \| \, \pi_{\theta}(\cdot \mid \mathcal{P}_{\text{wk}}(I), q, o_{<t}) \big)
\end{equation}

$S'_t$ probes the local curvature of the policy distribution on the perceptual manifold. It functions as a stability probe to identify numerical instability rather than semantic shift. A robust \textit{pivotal perceptual tokens} must exhibit high $S_t$ (strong \textit{Visual Dependency}) but low $S'_t$ (weak \textit{Visual Inconsistency}). 
% A high $S'_t$ relative to $S_t$ unmasks "brittle" tokens where the model's confidence is driven by high-frequency pixel fluctuations rather than invariant visual information.

\begin{figure*}[!ht]
    \centering
    % 第一张图
    \begin{minipage}[b]{0.57\textwidth}
        \centering
        \includegraphics[width=\textwidth]{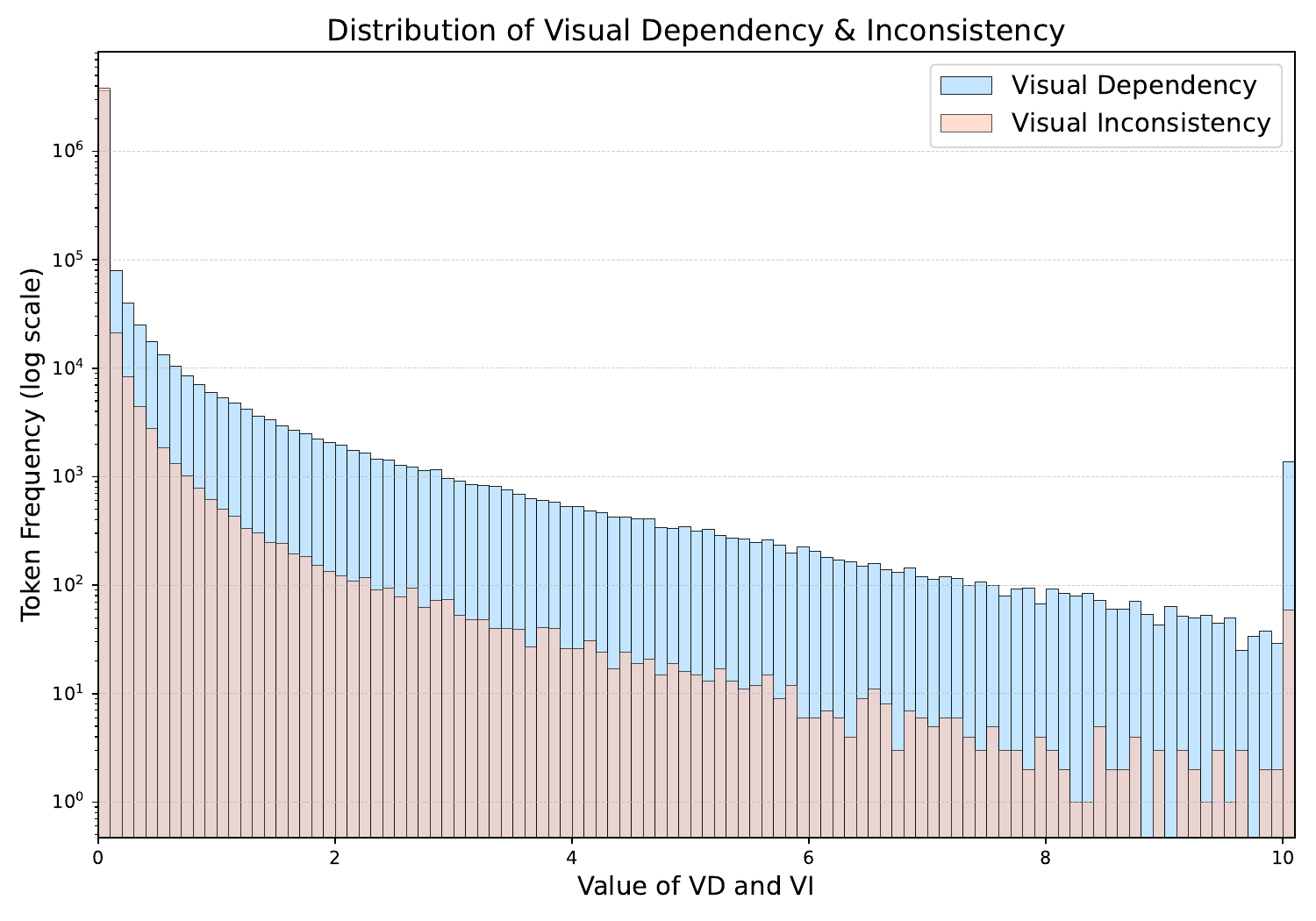}
        \caption{Distribution of Visual Dependency (VD) and Visual Inconsistency (VI) on Qwen2.5-VL-7B.}
        \label{fig:distribution}
    \end{minipage}
    \hfill
    % 第二张图
    \begin{minipage}[b]{0.40\textwidth}
        \centering
        \includegraphics[width=\textwidth]{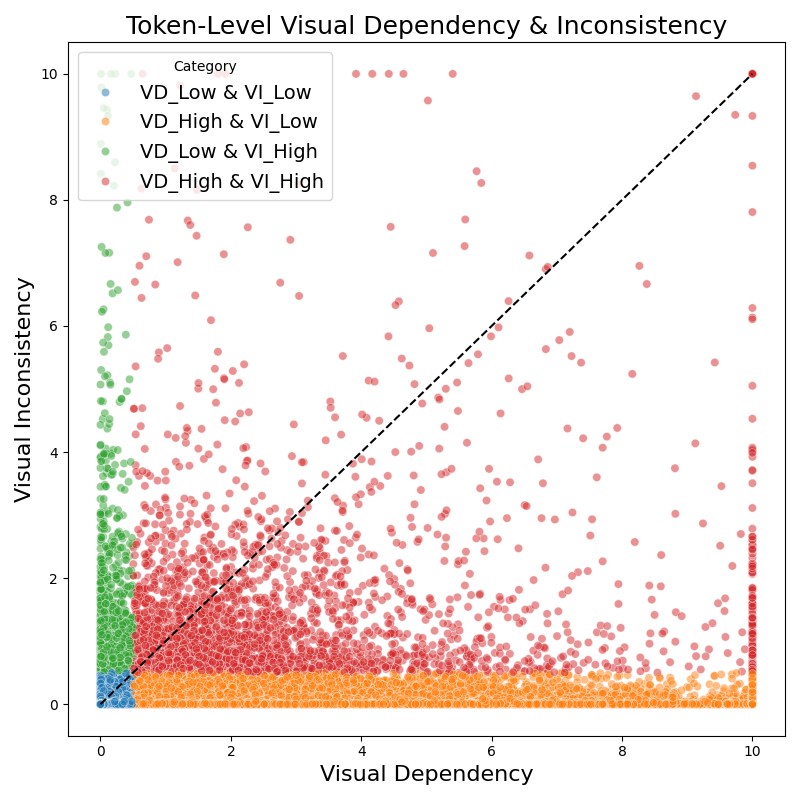}
        \caption{Token-Level distribution of VD and VI on Qwen2.5-VL-7B.}
        \label{fig:distribution_token}
    \end{minipage}
\end{figure*}

To provide a principled foundation for PRPO, we conduct a fine-grained diagnostic analysis on the MathVerse benchmark using Qwen2.5-VL-7B with a rollout factor of 8. By computing $S_t$ and $S'_t$ for each tokens, we uncover two critical findings regarding the distribution of visual signals:

\paragraph{Finding 4.1: Visual Signal Sparsity.} 
Our analysis reveals that perceptual tokens are extremely sparse within long-horizon reasoning trajectories. As illustrated in Figure~\ref{fig:distribution}, both $S_t$ and $S'_t$ exhibit a dramatic \textbf{long-tail characteristic}. The vast majority of generated tokens reside in the low-value region ($S_t, S'_t < 0.1$), suggesting that the policy is predominantly driven by language priors rather than visual evidence. Traditional RL methods (e.g., GRPO and DAPO) that broadcast advantages uniformly across the trajectory suffer from \textbf{gradient dilution} in multi-modal context, where rare but critical visual signals are submerged by linguistic background noise.

\paragraph{Finding 4.2: The Coupled Nature of Visual Dependency \& Inconsistency.}
Visual Dependency ($S_t$) and Inconsistency ($S'_t$) are inherently coupled rather than trivially separable. By projecting tokens across both dimensions (Figure~\ref{fig:distribution_token}), we can roughly partition them into three distinct regimes:

\begin{itemize}[leftmargin=*]
% , nosep

\item \textbf{Linguistic Tokens (low $S_t$).}
The vast majority of tokens fall into this region (the blue cluster). Unaffected by either perturbation, they are governed entirely by linguistic priors.
% or reasoning templates rather than visual evidence.
% This observation explains the extreme sparsity of perceptual signals observed in Finding 4.1.

\item \textbf{Brittle Anchors (high $S_t$, high $S'_t$).}
A substantial subset of tokens exhibits strong visual dependency but equally high sensitivity to numerically unstable pixel-level features (the red cluster). 
Reinforcing them solely based on $S_t$ would amplify noise rather than informative perception.

\item \textbf{Pivotal Perceptual Tokens (high $S_t$, low $S'_t$).}
The true \textit{pivotal perceptual tokens} form a narrow and highly structured subset (the orange cluster). 
These tokens exhibit strong semantic grounding while remaining stable under local perturbations, indicating that their predictions are driven by informative visual evidence. 
We refer to these tokens as \textit{pivotal perceptual tokens}, as they represent the causal interface where visual evidence is incorporated into the reasoning trajectory.

\end{itemize}

This decomposition exposes a fundamental limitation of pivotal token selection strategies, such as Jensen-Shannon Divergence, single KL Divergence~\citep{huang2025spotlight} and hidden state similarity~\citep{li2026rethinking}, which fail to distinguish \textbf{semantically grounded perception} from \textbf{numerically brittle responses}. 
A dual-metric synergistic approach ($S_t \uparrow, S'_t \downarrow$) is necessary to filter out stable \textit{pivotal perceptual tokens} that drive visually-grounded reasoning. Further performance analysis of these metrics is provided in the Appendix~\ref{app:heuristic_metrics}.

\paragraph{Definition 4.3 (Robust Visual Dependency).} 
\textit{Let $S_t$ and $S'_t$ be the Visual Dependency and Visual Inconsistency at step of token $t$, respectively. The hyperparameter $\eta$ ensures numerical stability and serves as a baseline sensitivity threshold. We define the Robust Visual Dependency (}$\text{RVD}_t$\textit{) to precisely quantify the real contribution of visual perception to the generation of token $o_t$:} 
% isolate pivotal perceptual token from the reasoning trajectory.}

\begin{equation}
\text{RVD}_t = S_t \cdot \text{ReLU} \left(1 - \frac{S'_t}{S_t + \eta} \right)
\label{eq:rvd}
\end{equation}

The RVD index acts as a non-linear information filter. By utilizing the ReLU-gated ratio, it effectively suppresses tokens in the \textbf{stochastically brittle} region (where $S'_t \geq S_t$) while preserving the original magnitude of \textbf{Golden Anchors} (where $S'_t \ll S_t$). This ensures that the subsequent reinforcement signal is concentrated on tokens that are both semantically grounded and numerically robust.

\subsection{Token-Level Dynamic Perceptual Advantage Reshaping (PAR)}

While RVD provides a fine-grained measure of perceptual relevance, directly using it for credit assignment leads to suboptimal optimization behavior due to its highly skewed distribution:

\textbf{(1) Long-tailed distribution.}
RVD exhibits a heavy-tailed structure, where the majority of tokens concentrate near zero, while only a small subset of perceptually grounded tokens take large values. Directly using such signals results in inefficient credit allocation.

\textbf{(2) Gradient collapse.}
A naive transformation (e.g., linear or monotonic scaling) causes most token-level weights to shrink toward zero, effectively collapsing gradients onto a sparse subset of visual tokens. This leads to a biased optimization process that over-focuses on perceptual tokens while neglecting linguistic reasoning, ultimately degrading general reasoning performance.

% To tackle these problems, we propose \textbf{Perceptual Advantage Reshaping (PAR)}. Inspired by the activation dynamics of biological neurons, PAR functions as a differentiable synaptic gating mechanism. It maps raw perceptual stimuli into an \emph{S-shaped synaptic response curve}, ensuring a bio-plausible phase transition where weak stimuli are suppressed, critical stimuli trigger rapid neuronal firing, and excessive signals smoothly saturate.

To address these issues, we propose \textbf{Perceptual Advantage Reshaping (PAR)}, a differentiable nonlinear mapping that transforms raw perceptual signals into bounded token-level weights. The resulting \textit{S-shaped synaptic response curve} suppresses low-signal tokens, amplifies informative perceptual tokens beyond a threshold, and enforces smooth saturation, thereby preventing gradient explosion while preserving stable optimization.

\paragraph{Synaptic S-shaped Reshaping.}
Formally, let $x = \mathrm{RVD}_t \in \mathbb{R}^+$ denote the raw perceptual stimulus. To achieve a strictly bounded and mathematically smooth transformation, we formulate the reshaped advantage $\mathcal{T}(x)$ via a generalized logistic function:

\begin{equation}
\mathcal{T}(x) = m + \frac{M - m}{1 + \exp\left(-k \cdot (x - \mu)\right)}
\end{equation}

where the synaptic activation is governed by four highly interpretable parameters: the activation threshold $\mu > 0$, the upper saturation bound $M$ and the lower suppression baseline $m$ ($M > m \ge 0$), and a scaling factor $k > 0$ that dictates the sharpness of the activation.

% The induced transformation $\mathcal{T}(\cdot)$ organically divides the signal space into three functional regimes, directly mapping neurobiological phenomena to desirable inductive biases for the network.
% :

% \begin{itemize}[leftmargin=*]
%     \item \textbf{Sub-threshold Suppression ($x \ll \mu$).} 
%     Analogous to a neuron's resting state, low-RVD signals are heavily attenuated toward the baseline $m$. This soft sparsity effectively filters out perceptual noise while preserving a marginal gradient flow to prevent dead neurons.
%     \item \textbf{Neuronal Firing Transition ($x \approx \mu$).}
%     In the neighborhood of the threshold, the exponential term drives a rapid, near-linear amplification. This mimics the generation of an action potential (neuronal firing), selectively routing intense gradients to amplify tokens that successfully cross the perceptual threshold.
%     \item \textbf{Synaptic Saturation ($x \gg \mu$).}
%     As the stimulus intensifies, the activation asymptotically saturates at $M$. This bounded supremum mirrors the refractory period of biological synapses, serving as an inherent regularizer to prevent extreme perceptual outliers from dominating the attention mechanism and destabilizing the forward pass.
% \end{itemize}

\paragraph{Token-level Dynamic Perceptual Advantage Reshaping (PAR).}
Standard trajectory-level reinforcement learning suffers from coarse credit assignment, uniformly broadcasting a scalar reward to all tokens regardless of their perceptual contributions. To resolve this, we leverage our synaptic response function $\mathcal{T}(\cdot)$ to establish \textbf{PAR}:

\begin{equation}
\hat{A}_{t}^{i,\text{PAR}} = \hat{A}^i \cdot \mathcal{T}(\mathrm{RVD}_t^i)
\end{equation}

This mechanism organically redistributes the learning signals. Tokens exhibiting strong perceptual relevance (post-threshold) receive amplified advantages, directing the optimization focus toward semantically pivotal regions. Conversely, perceptual noise is softly discounted. Critically, the strictly positive lower bound $m$ in $\mathcal{T}(\cdot)$ guarantees a non-zero gradient even for suppressed tokens, preserving dense learning signals and preventing optimization dead zones across the sequence.

\paragraph{Training Objectives.}
We formulate our final policy objective by integrating \textbf{PAR} into a clipped surrogate framework for robust optimization:

\begin{equation}
\label{training_objective}
\mathcal{J}_{\text{PRPO}}(\theta) = \mathbb{E} \Bigg[
\frac{1}{\sum_{i=1}^{G} |o^i|} \sum_{i=1}^{G} \sum_{t=1}^{|o^i|} 
\min \Big( 
r_{t}^i(\theta) \hat{A}_{t}^{i,\text{PAR}}, \;
\text{clip}\big(r_{t}^i(\theta), 1 - \epsilon_{\text{low}}, 1 + \epsilon_{\text{high}} \big) \hat{A}_{t}^{i,\text{PAR}} 
\Big) 
\Bigg]
\end{equation}

By bounding dynamically reshaped advantages within an asymmetric trust region, PRPO achieves precise multimodal alignment without overfitting to visual noise or compromising language priors.

\section{Experiments}

\begin{table*}[!ht]
\renewcommand{\arraystretch}{0.8}
\centering
\caption{Main results ($\text{Rollout} = 8$) on various benchmarks. The best results are \textbf{bolded}, and the second-best results are \underline{underlined}. Comparison is performed separately within 3B and 7B groups.}
\label{tab:main_results}
\small
\setlength{\tabcolsep}{1.5pt}
\begin{tabular}{l|cccccccc|c}
\toprule
\multicolumn{1}{c|}{Model} & Geo3k & MathVista & We-Math & MMK12 & MathVerse & Logic & MMMU-Pro & $\text{MathVerse}_{V}$ & AVG \\
\midrule
\multicolumn{10}{c}{\textit{3B Models}}\\
\midrule
\rowcolor{gray!20}Qwen2.5-VL-3B & 18.8 & 46.4 & 29.8 & 37.1 & 33.8 & 30.8 & 19.3 & 30.1 & 30.7 \\
\quad + GRPO & 28.7 & 59.3 & 58.9 & 57.2 & 55.3 & 38.1 & 25.7 & 52.3 & 46.9 \\
\quad + DAPO & 31.2 & 60.9 & 60.0 & \underline{66.8} & 56.3 & 40.7 & 28.4 & 53.1 & 49.7 \\
\quad + PAPO-G & 31.1 & 59.7 & 59.9 & 56.3 & 57.5 & 39.5 & 26.4 & 53.6 & 48.0 \\
\quad + PAPO-D & 35.7 & 62.5 & \underline{62.7} & 64.1 & 60.5 & 41.7 & 28.8 & 57.7 & 51.7 \\
\quad + VPPO & \underline{35.8} & \underline{63.6} & 61.4 & 65.3 & \underline{61.3} & \underline{42.0} & \underline{28.8} & \underline{58.1} & \underline{52.1} \\
\rowcolor{green!10}\quad + PRPO(Ours) & \textbf{37.3} & \textbf{65.3} & \textbf{63.9} & \textbf{66.9} & \textbf{62.7} & \textbf{44.9} & \textbf{30.3} & \textbf{60.7} & \textbf{54.0} \\
\midrule
\multicolumn{10}{c}{\textit{7B Models}}\\
\midrule
MM-Eureka-7B & 40.3 & 71.9 & 66.1 & 67.7 & 67.2 & 46.3 & 30.3 & 62.4 & 56.5 \\
VL-Rethinker-7B & 40.7 & \underline{70.6} & 69.3 & 68.3 & 68.8 & 46.5 & 37.1 & 65.0 & 58.6 \\
NoisyRollout-7B & \textbf{52.0} & 68.6 & \underline{69.6} & 51.0 & 67.8 & 47.5 & 34.5 & 63.8 & 56.8 \\
R1-ShareVL-7B & 41.2 & 69.3 & 68.7 & 70.9 & 68.2 & 45.8 & 35.1 & 64.3 & 57.9 \\

\rowcolor{gray!20}Qwen2.5-VL-7B & 36.7 & 63.6 & 46.7 & 43.3 & 39.5 & 42.0 & 25.2 & 34.3 & 41.4 \\
\quad + GRPO & 40.2 & 65.5 & 68.1 & 72.3 & 66.5 & 45.6 & 35.2 & 61.7 & 56.9 \\
\quad + DAPO & 35.9 & 61.9 & 58.5 & 75.9 & 55.6 & 37.1 & 29.0 & 51.0 & 50.6 \\
\quad + PAPO-G & 40.3 & 68.1 & 66.8 & 71.8 & 68.3 & 45.8 & 35.5 & 64.1 & 57.6 \\
\quad + PAPO-D & 44.1 & 67.5 & 68.3 & \underline{80.6} & 68.6 & 46.7 & 36.3 & 64.9 & 59.6 \\
\quad + VPPO & 44.4 & 70.5 & 68.9 & 80.1 & \underline{71.0} & \underline{47.5} & \underline{38.8} & \underline{65.5} & \underline{60.8} \\
\rowcolor{green!10}\quad + PRPO(Ours) & \underline{46.9} & \textbf{72.2} & \textbf{70.8} & \textbf{82.8} & \textbf{72.2} & \textbf{49.7} & \textbf{38.8} & \textbf{66.3} & \textbf{62.5} \\
\bottomrule
\end{tabular}
\end{table*}
\subsection{Experimental Setup}

\paragraph{Baseline Models, Data and Methods.}
To ensure a rigorous and equitable comparison with prior SOTA approaches, we deliberately align our experimental setup—including backbone architectures and training corpora-with established conventions in recent literature. Specifically, we instantiate our PRPO framework utilizing the Qwen2.5-VL series (both 3B and 7B variants) as the base policy models, and conduct training exclusively on the \texttt{ViRL39K}~\citep{wang2025vl} dataset. Furthermore, we benchmark PRPO against nine prominent training-based methods, encompassing both foundational reinforcement learning paradigms and recent multimodal-specific advancements. These baselines include GRPO~\citep{shao2024deepseekmath}, DAPO~\citep{yu2025dapo}, PAPO~\citep{wang2025perception}, VPPO~\citep{huang2025spotlight}, MM-Eureka-7B~\citep{meng2025mm}, VL-Rethinker-7B~\citep{wang2025vl}, NoisyRollout-7B~\citep{liu2025noisyrollout}, and R1-ShareVL-7B~\citep{yao2025r1}.

\paragraph{Evaluation Benchmarks.} For a comprehensive evaluation of model capabilities, we assess performance across seven widely adopted benchmarks spanning complex mathematical reasoning and general multimodal understanding: Geo3K~\citep{lu2021inter}, MathVista~\citep{lu2023mathvista}, We-Math~\citep{qiao2024we}, MMK12~\citep{meng2025mm}, MathVerse~\citep{zhang2024mathverse}, LogicVista~\citep{xiao2024logicvista}, MMMU-Pro~\citep{yue2024mmmu}. $\text{MathVerse}_{V}$ denotes the vision-centric subset of MathVerse.

\paragraph{Training Details.} 
To ensure a rigorous and equitable comparison with baseline training methodologies, we align our experimental protocol with established settings of prior works. Specifically, both the 3B and 7B models are trained on the \texttt{ViRL39K}~\citep{wang2025vl} dataset for 2 epochs, employing a learning rate of 1e-6 and a rollout batch size of 384. The maximum response length is constrained to 2048 tokens. Comprehensive hyperparameter settings are detailed in Appendix~\ref{recipe}.

\subsection{Experimental Results}

As presented in Table \ref{tab:main_results}, our proposed PRPO establishes SOTA performance across both 3B and 7B models, demonstrating comprehensive superiority over existing multimodal RL algorithms. 

\textbf{SOTA Performance and Consistency.} 
PRPO achieves the best average performance on both the 3B and 7B models. Compared to the baseline Qwen2.5-VL Series, our models deliver substantial improvements of +23.3\% (from 30.7\% to 54.0\%) and +21.1\% (from 41.4\% to 62.5\%) at the 3B and 7B scales, respectively. Compared to the strongest baseline method VPPO, our models exhibit
\begin{wrapfigure}[12]{l}{0.31\textwidth}
% \vspace{-5mm}
  \centering
  \includegraphics[width=0.31\textwidth]{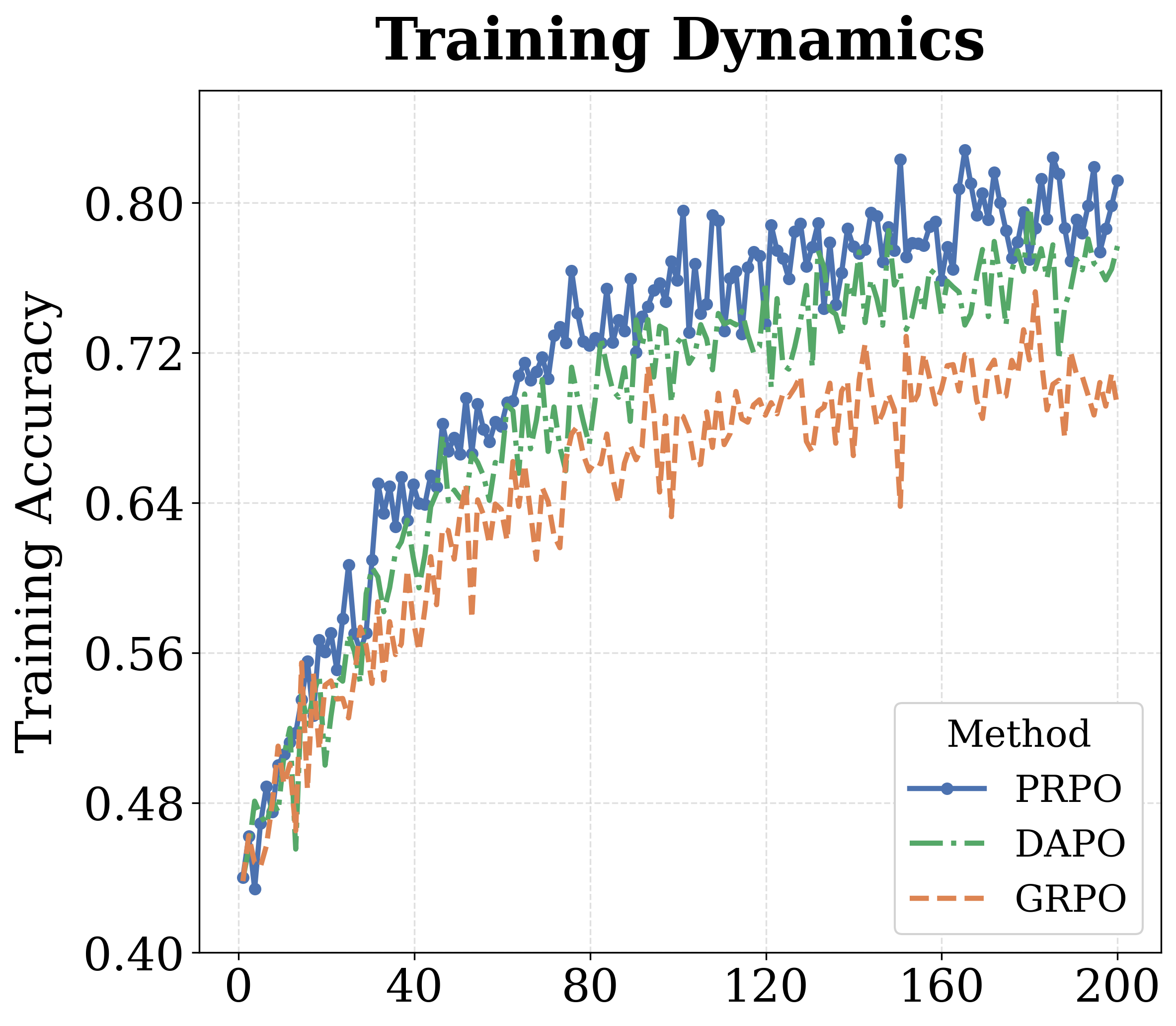}
    \caption{Training dynamics.}
  \label{fig:training}
\end{wrapfigure}
improvements of +1.9\% and +1.7\%. Furthermore, we achieve consistent performance gains across multiple tasks, demonstrating that PRPO significantly outperforms existing training methods.

\textbf{Dynamic Token-level Credit Assignment.} 
A key distinction between PRPO and high-performing baselines like PAPO and VPPO lies in the granularity of credit assignment. While existing methods primarily rely on coarse \textbf{trajectory-level} reward signals, PRPO introduces a \textbf{dynamic token-level} advantage reshaping framework. This empirical gap validates that our proposed \textbf{RVD} and \textbf{PAR} techniques not only accurately identify \textit{perceptual pivotal tokens} on line, but also effectively amplifying those tokens while suppressing irrelevant noise of language priors.

% \vspace{-7mm}
\textbf{Superior Training Efficiency.} 
As shown in Figure~\ref{fig:training}, PRPO exhibits significantly higher training efficiency than the GRPO and DAPO baselines. Under identical training step constraints, PRPO achieves markedly higher performance gains and exhibits faster convergence. This evidence demonstrates that the synaptic-inspired response curve provides high-quality, informative training signals, making PRPO a more efficient and scalable objective for the complex optimization landscape of LVLMs. See Appendix~\ref{sec:equal_budget} for additional training efficiency analysis.
% \vspace{14mm}
\subsection{Ablation Study}
\begin{table*}[!ht]
\centering
\renewcommand{\arraystretch}{0.8}
\caption{Ablation study of core components on Qwen2.5-VL-7B ($\text{Rollout} = 8$). The best results are \textbf{bolded}, and the second-best results are \underline{underlined}.}
\label{tab:ablation_method}
\small
\setlength{\tabcolsep}{1pt}
\begin{tabular}{l|cccccccc|c}
\toprule
\multicolumn{1}{c|}{Model} & Geo3k & MathVista & We-Math & MMK12 & MathVerse & Logic & MMMU-Pro & $\text{MathVerse}_{V}$ & AVG \\
\midrule
\rowcolor{gray!20}Qwen2.5-VL-7B & 36.7 & 63.6 & 46.7 & 43.3 & 39.5 & 42.0 & 25.2 & 34.3 & 41.4 \\
\midrule
\quad + DAPO & 35.9 & 61.9 & 58.5 & 75.9 & 55.6 & 37.1 & 29.0 & 51.0 & 50.6 \\
\quad + PRPO w/o VI & \underline{44.8} & \underline{71.0} & \underline{69.2} & \underline{80.5} & \underline{71.3} & \underline{48.1} & \underline{38.6} & \underline{65.8} & \underline{61.2} \\
\quad + PRPO w/o PAR  & 43.9 & 69.9 & 68.4 & 79.5 & 70.2 & 46.9 & 38.1 & 65.0 & 60.2 \\
\midrule
\rowcolor{green!10}\quad + PRPO (Ours) & \textbf{46.9} & \textbf{72.2} & \textbf{70.8} & \textbf{82.8} & \textbf{72.2} & \textbf{49.7} & \textbf{38.8} & \textbf{66.3} & \textbf{62.5} \\
\bottomrule
\end{tabular}
\end{table*}
% \vspace{-5mm}
\paragraph{Ablation Study of Core Components.}
Table \ref{tab:ablation_method} demonstrates that both the Visual Inconsistency (VI) penalty and Perceptual Advantage Reshaping (PAR) are essential for multimodal RL. Removing the VI module (w/o VI) exposes the policy to single Visual Dependencies (VD), which increases hallucination in long reasoning chains (AVG drops to 61.2\%). More importantly, removing the PAR mechanism (w/o PAR) causes a severe performance degradation to 60.2\%. Without the bounded non-linear gating, large gradients from perceptual tokens dominate the linguistic priors, leading to representation collapse. Notably, even these ablated variants significantly outperform the trajectory-level DAPO baseline (50.6\%), empirically demonstrating that fine-grained token-level routing is a fundamentally superior paradigm for multimodal RL.

\begin{table*}
\centering
\caption{Ablation study of key hyperparameters in Perceptual Advantage Reshaping (PAR) on Qwen2.5-VL-7B ($\text{Rollout} = 8$). The best results are \textbf{bolded}.}
\label{tab:ablation_hyper}
\renewcommand{\arraystretch}{0.7}
\small
\setlength{\tabcolsep}{4.2pt}
\begin{tabular}{ccc|ccccccc|c}
\toprule
$M$ & $m$ & $k$ & Geo3k & MathVista & We-Math & MMK12 & MathVerse & Logic & MMMU-Pro & AVG \\
\midrule
\rowcolor{gray!20}\multicolumn{3}{c|}{\textit{Baseline}} & 36.7 & 63.6 & 46.7 & 43.3 & 39.5 & 42.0 & 25.2 & 42.4 \\
\midrule
% Ablation on Upper Saturation Bound (M)
\multicolumn{11}{c}{\textit{Ablation on Upper Saturation Bound ($M$)}}\\
\midrule
2.0 & 0.10 & 2.0 & 43.2 & 68.5 & 65.2 & 79.5 & 67.1 & 46.1 & 35.8 & 57.9 \\
6.0 & 0.10 & 2.0 & 44.5 & 69.8 & 67.5 & 78.0 & 68.5 & 47.2 & 36.5 & 58.9 \\
\midrule
% Ablation on Lower Suppression Baseline (m)
\multicolumn{11}{c}{\textit{Ablation on Lower Suppression Baseline ($m$)}}\\
\midrule
4.0 & 0.05 & 2.0 & 45.1 & 70.2 & 68.1 & 80.5 & 70.1 & 48.9 & 37.2 & 60.0 \\
4.0 & 0.15 & 2.0 & 44.8 & 69.5 & 66.8 & 77.2 & 68.9 & 45.5 & 36.4 & 58.4 \\
\midrule
% Ablation on Steepness (k)
\multicolumn{11}{c}{\textit{Ablation on Activation Sharpness ($k$)}}\\
\midrule
4.0 & 0.10 & 1.5 & 43.9 & 68.8 & 67.1 & 78.4 & 69.2 & 46.8 & 36.1 & 58.6 \\
4.0 & 0.10 & 2.5 & 45.5 & 71.0 & 69.2 & 81.1 & 70.8 & 47.5 & 37.8 & 60.4 \\
\midrule
% Optimal Configuration (Ours)
\rowcolor{green!10}4.0 & 0.10 & 2.0 & \textbf{46.9} & \textbf{72.2} & \textbf{70.8} & \textbf{82.8} & \textbf{72.2} & \textbf{49.7} & \textbf{38.8} & \textbf{61.9} \\
\bottomrule
\end{tabular}
\end{table*}

\paragraph{Ablation Study of PAR Curve Parameters.}
We further investigate the sensitivity of synaptic gating function which is mainly shaped by $(M, m, k)$ to ensure mathematical stability in optimization landscape. As shown in Table~\ref{tab:ablation_hyper}, we can summarize the specific effects of these parameters as follows:

\begin{itemize}[leftmargin=*, topsep=2pt, itemsep=2pt]
    \item \textbf{Upper Saturation Bound ($M$):} A conservative bound ($M=2.0$) inadequately amplifies critical tokens (AVG: 57.9\%). Conversely, an excessive bound ($M=6.0$) triggers reward dilution and high-variance updates. $M=4.0$ optimally balances signal amplification and gradient regularization.
    
    \item \textbf{Lower Suppression Baseline ($m$):} This soft lower bound acts as a critical inductive bias against optimization dead zones. Over-suppressing it ($m=0.05$) starves the gradient flow for subtle visual contexts, whereas elevating it ($m=0.15$) paralyzes the noise-filtering capacity.
    
    \item \textbf{Activation Sharpness ($k$):} Modulating the phase transition of the gating mechanism, a dampened curve ($k=1.5$) homogenizes the advantage distribution, failing to isolate critical tokens. In contrast, an aggressive step-like transition ($k=2.5$) destabilizes optimization near the threshold. Setting $k=2.0$ guarantees a smooth, robust landscape for optimal credit assignment.
\end{itemize}

\begin{table*}
\centering
\caption{Ablation study on the activation threshold ($\mu$) percentile of RVD on Qwen2.5-VL-7B ($\text{Rollout} = 8$). The best results are \textbf{bolded}.}
\label{tab:ablation_threshold}
\renewcommand{\arraystretch}{0.8}
\small
\setlength{\tabcolsep}{4pt}
\begin{tabular}{c|ccccccc|c}
\toprule
Threshold ($\mu$) & Geo3k & MathVista & We-Math & MMK12 & MathVerse & Logic & MMMU-Pro & AVG \\
\midrule
\rowcolor{gray!20}\textit{Baseline} & 36.7 & 63.6 & 46.7 & 43.3 & 39.5 & 42.0 & 25.2 & 42.4 \\
\midrule
20th Percentile & 46.1 & 71.5 & 69.5 & 81.6 & 71.4 & 48.8 & 38.0 & 61.0 \\
\rowcolor{green!10}25th Percentile & \textbf{46.9} & \textbf{72.2} & \textbf{70.8} & \textbf{82.8} & \textbf{72.2} & \textbf{49.7} & \textbf{38.8} & \textbf{61.9} \\
30th Percentile & 45.2 & 70.4 & 68.3 & 80.2 & 70.1 & 47.9 & 37.3 & 59.9 \\
40th Percentile & 39.2 & 65.8 & 52.4 & 55.1 & 45.6 & 43.5 & 28.6 & 47.2 \\
\bottomrule
\end{tabular}
\end{table*}

\vspace{-5mm}
\paragraph{Ablation Study of the Activation Threshold ($\mu$).}
As shown in Table~\ref{tab:ablation_threshold}, we observe that setting the threshold $\mu$ to \textbf{the 25th percentile} of RVD yields the best performance. This indicates that reinforcing only the \textbf{Top-25\%} of \textit{pivotal perceptual tokens} is sufficient to achieve effective perceptual optimization, while maintaining a lower gradient level for the remaining 75\% of tokens effectively filters linguistic noise and ensures the steady evolution of foundational reasoning capabilities.

\subsection{Case Study: More Accurate Reasoning with Fewer Tokens}
\vspace{-3mm}
\begin{figure*}[!ht]
\centering
  \includegraphics[width=1.0\linewidth]{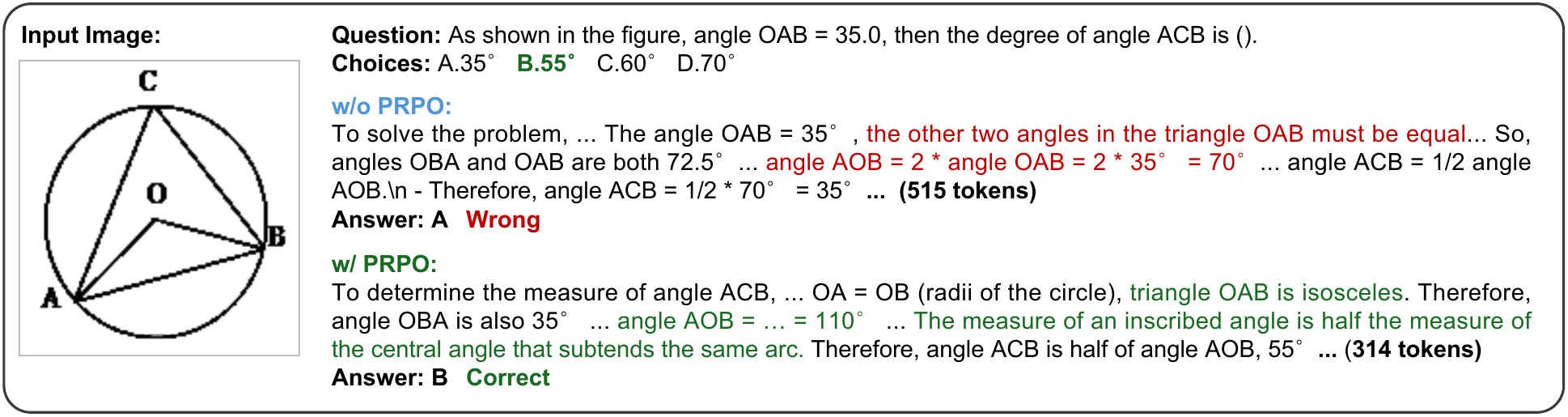}
  \caption{Comparison of the generation results between PRPO the baseline using Qwen2.5-VL-7B. }
  \label{case}
\end{figure*}
\vspace{-3mm}
As shown in Figure~\ref{case}, PRPO successfully corrects the perception errors of the baseline. Consequently, the entire reasoning trajectory becomes much more concise, significantly reducing required tokens.

\section{Conclusion}

We propose PRPO, a token-level reinforcement learning framework that addresses the Perceptual Credit Assignment Problem in multimodal reasoning. By introducing Robust Visual Dependency (RVD) and Perceptual Advantage Reshaping (PAR), we move beyond coarse trajectory-level rewards to precisely reinforce pivotal perceptual tokens while filtering out numerical noise and linguistic priors. Our framework achieves SOTA performance across seven benchmarks, with average gains of 23.3\% and 21.1\% at 3B and 7B scales, respectively. These results demonstrate that fine-grained token-level credit assignment is essential for aligning multimodal reasoning systems at scale.

\bibliographystyle{plainnat}
\bibliography{neurips_2026}

\clearpage

\appendix

\section{Experimental Details}

\subsection{Benchmarks and Evaluation Metrics}

To provide a comprehensive evaluation and comparison of different LVLM RL methods, we align with prior literature and select the following benchmarks across diverse difficulty levels and domains:

\textbf{(1) Geometric and Mathematical Reasoning.} The primary focus of our experiments lies in assessing the model's ability to navigate various mathematical and geometric domains.
\begin{itemize}[leftmargin=*]
    \item \textbf{Geo3k}~\citep{lu2021inter}: This benchmark specializes in geometry-centric problems at the high school level. Its rigorous use of formal language annotations allows for a precise evaluation of how well models can execute symbolic reasoning within a visual context.
    \item \textbf{MathVista}: Serving as a comprehensive evaluation of visually situated mathematics, this benchmark integrates a broad spectrum of visual contexts—ranging from statistical charts and functional plots to abstract geometry and real-world scenes. It systematically challenges models to synthesize complex visual evidence with deep quantitative and logical deduction, providing a robust measure of generalized mathematical capability across highly diverse visual modalities.
    \item \textbf{We-Math}~\citep{qiao2024we}: Focused on a human-centric diagnostic approach, We-Math breaks down complex problems into a hierarchy of 67 specific knowledge points. This granularity allows us to identify whether a failure stems from a lack of specific knowledge or a breakdown in the reasoning chain.
    \item \textbf{MMK12}~\citep{meng2025mm}: This benchmark provides a standardized assessment of foundational math skills derived from K-12 curricula. It ensures that the model possesses the prerequisite quantitative literacy required for more specialized tasks.
    \item \textbf{MathVerse}~\citep{zhang2024mathverse}: This benchmark probes the visual dependency of Large Vision-Language Models. By offering six variants of the same problem with differing ratios of visual-to-textual information, it uncovers whether a model truly "perceives" a diagram or simply exploits linguistic redundancies.
\end{itemize}

\textbf{(2) Logical and Cognitive Foundations.} Beyond pure mathematics, we evaluate the model's underlying cognitive architecture through targeted logical tasks.
\begin{itemize}[leftmargin=*]
    \item \textbf{LogicVista}~\citep{xiao2024logicvista}: Designed to measure general-purpose logic, this benchmark transcends the mathematical domain by testing five core reasoning pillars: induction, deduction, and spatial, mechanical, and numerical reasoning. It offers a diverse array of visual formats to assess fundamental intellectual agility.
\end{itemize}

\textbf{(3) Advanced Multimodal Integration}. To measure the upper limits of university-level understanding where visual and textual data must be processed as a single entity, we utilize a specialized "pro" suite.
\begin{itemize}[leftmargin=*]
    \item \textbf{MMMU-Pro}~\citep{yue2024mmmu}: As a more rigorous iteration of the MMMU dataset, this version is specifically engineered to defeat text-only shortcuts. By increasing the number of distractors and incorporating vision-exclusive formats, it demands that the model perform high-level academic reasoning that is fundamentally impossible without deep multimodal fusion.
\end{itemize}

\subsection{Baseline Methods}
Recently, researchers have optimized reinforcement learning for LVLMs from various perspectives. Specifically, we compare PRPO with the following closely related baseline methods:

\begin{itemize}[leftmargin=*]
    \item \textbf{PAPO}~\citep{wang2025perception} introduces a policy gradient algorithm that encourages reasoning steps to be internally grounded in visual inputs without relying on external annotations or reward models. However, its optimization objective is applied uniformly across all generated tokens, neglecting the inherently sparse nature of critical visual signals.
    \item \textbf{VPPO}~\citep{huang2025spotlight} attempts to address the uniform signal bottleneck through a two-tiered strategy, employing trajectory-level reward shaping combined with focusing policy updates on a sparse set of perception tokens.
\end{itemize}

\textbf{Differences from our work:} While both baselines seek to enhance visually-grounded reasoning, they exhibit key limitations that PRPO explicitly resolves. PAPO's uniform token treatment inevitably leads to gradient dilution, where critical visual signals are overwhelmed by abundant language priors. VPPO mitigates this by focusing on sparse tokens, but its reliance on coarse trajectory-level reward shaping and single-metric token selection risks amplifying numerically brittle visual noise. 

In contrast, our PRPO framework operates through a highly precise, dynamic \textit{token-level} paradigm. By jointly evaluating Visual Dependency and Visual Inconsistency via the Robust Visual Dependency (RVD) metric, PRPO systematically filters out stochastic noise to isolate genuinely pivotal perceptual tokens. Furthermore, instead of discrete token masking or trajectory-level adjustments, PRPO utilizes Perceptual Advantage Reshaping (PAR)---a continuous, synaptic S-shaped gating mechanism that dynamically amplifies critical visual signals while preserving stable, non-zero optimization gradients for foundational reasoning capabilities.
\subsection{Training Recipe}
\label{recipe}
To ensure a fair and rigorous comparison, we adopt the same foundational experimental configurations as the aforementioned PAPO and VPPO baselines. Our modifications are strictly confined to the introduction of specific hyperparameters governing the token-level advantage reshaping. The comprehensive training parameters are detailed in Table \ref{tab:hyperparameters}.

\begin{table}[!ht]
    \centering
    \caption{Key hyperparameters for training and evaluation.}
    \label{tab:hyperparameters}
    \small
    \begin{tabular}{ll}
        \toprule
        \textbf{Hyperparameter} & \textbf{Value} \\
        \midrule
        \multicolumn{2}{l}{\textit{General Training}} \\
        Optimizer & AdamW \\
        Learning Rate & 1e-6 \\
        LR Schedule & Constant (no warmup or decay) \\
        Epochs & 2 \\
        Freeze Vision Tower & False \\
        \midrule
        \multicolumn{2}{l}{\textit{RL Process}} \\
        Global Batch Size & 128 \\
        Rollout Batch Size & 384 \\
        Rollouts per Prompt & 8 \\
        Rollout Top-p & 0.99 \\
        Max Response Length & 2048  \\
        Reward Signal & Binary Accuracy (1/0) \\
        \midrule
        \multicolumn{2}{l}{\textit{DAPO Recipe}} \\
        Sampling Method & Dynamic Sampling \\
        Clip Ratio Low & 0.2 \\
        Clip Ratio High & 0.28 \\
        Loss Averaging Mode & Token-level \\
        KL Penalty & None \\
        \midrule
         \multicolumn{2}{l}{\textit{VPPO Specific}} \\
         TAS $\beta_{\min}$ & 0.9 \\
         TAS $\beta_{\max}$ & Dynamical (batch-normalized) \\
         TGF Ratio ($k$) & 0.4 \\
        \midrule
        \rowcolor{gray!10} \multicolumn{2}{l}{\textit{PRPO Specific (Ours)}} \\
        \rowcolor{gray!10} Activation Threshold ($\mu$) & 25th Percentile \\
        \rowcolor{gray!10} Upper Saturation Bound ($M$) & 4.0 \\
        \rowcolor{gray!10} Lower Suppression Baseline ($m$) & 0.1 \\
        \rowcolor{gray!10} Activation Sharpness ($k$) & 2.0 \\
        \midrule
        \multicolumn{2}{l}{\textit{Evaluation Generation}} \\
        Temperature & 1.0 \\
        Top-p & 1.0 \\
        Max New Tokens & 2048 \\
        \bottomrule
    \end{tabular}
\end{table}

\subsection{Compute Resources}
All experiments were implemented using an 8 $\times$ NVIDIA A100 80GB GPUs.

\subsection{Algorithm Pseudocode of PRPO}

\begin{algorithm}[!ht]
\caption{The Perception-Reinforced Policy Optimization (PRPO) Algorithm}
\label{alg:prpo}
\begin{algorithmic}[1]
\State \textbf{Input:} Current policy $\pi_\theta$, old policy $\pi_{\theta_{\text{old}}}$, batch of prompts $D = \{(I_j, q_j)\}_{j=1}^{B}$
\State \textbf{Hyperparameters:} Group size $G$, numerical stabilizer $\eta$, PAR synaptic parameters $(\mu, M, m, k)$

\Procedure{PRPO\_Training\_Step}{$\pi_\theta, \pi_{\theta_{\text{old}}}, D$}
    \State Initialize lists for trajectories $\mathcal{T} \leftarrow []$, original distributions $\mathcal{P} \leftarrow []$
    
    \Comment{\textbf{Phase 1: Data Generation (Rollouts)}}
    \For{each prompt $(I, q)$ in $D$}
        \For{$i = 1$ to $G$}
            \State Generate trajectory $\tau_i = (o_1, ..., o_T)$ using $\pi_{\theta_{\text{old}}}( \cdot | I, q)$
            \State Store original distributions $P_i = \{\pi_{\theta_{\text{old}}}(\cdot | s_t, I)\}_{t=1}^T$
            \State Append $\tau_i$ to $\mathcal{T}$ and $P_i$ to $\mathcal{P}$
        \EndFor
    \EndFor

    \Comment{\textbf{Phase 2: Robust Visual Dependency (RVD) Evaluation}}
    \State Initialize list for robust visual dependencies $\mathcal{RVD} \leftarrow []$
    \For{each trajectory $\tau_i$ and its distributions $P_i$ in $(\mathcal{T}, \mathcal{P})$}
        \State Let $(I, q)$ be the prompt for $\tau_i$
        \State Generate Information-Corrupting Image: $I_{str} \leftarrow \mathcal{P}_{str}(I)$ \Comment{e.g., black mask}
        \State Generate Information-Preserving Image: $I_{wk} \leftarrow \mathcal{P}_{wk}(I)$ \Comment{e.g., pixel jitter}
        \State Compute distributions $P^{str}_i = \{\pi_{\theta_{\text{old}}}(\cdot | s_t, I_{str})\}_{t=1}^T$
        \State Compute distributions $P^{wk}_i = \{\pi_{\theta_{\text{old}}}(\cdot | s_t, I_{wk})\}_{t=1}^T$
        
        \State Initialize token-level RVD scores $RVD_i \leftarrow []$
        \For{$t=1$ to $T$}
            \State $S_{i,t} \gets D_{\text{KL}}(P_{i,t} \parallel P^{str}_{i,t})$ \Comment{Visual Dependency}
            \State $S'_{i,t} \gets D_{\text{KL}}(P_{i,t} \parallel P^{wk}_{i,t})$ \Comment{Visual Inconsistency}
            \State $RVD_{i,t} \gets S_{i,t} \cdot \text{ReLU}\left(1 - \frac{S'_{i,t}}{S_{i,t} + \eta}\right)$ \Comment{Filter brittle anchors}
            \State Append $RVD_{i,t}$ to $RVD_i$
        \EndFor
        \State Append $RVD_i$ to $\mathcal{RVD}$
    \EndFor
    
    \Comment{\textbf{Phase 3: Token-Level Perceptual Advantage Reshaping (PAR)}}
    \State Compute outcome rewards $\{R_i\}_{i=1}^{|\mathcal{T}|}$ and standard group advantages $\{\hat{A}_i\}_{i=1}^{|\mathcal{T}|}$
    \State Initialize list for token-level reshaped advantages $\hat{\mathcal{A}}^{PAR} \leftarrow []$
    \For{each trajectory $\tau_i$, advantage $\hat{A}_i$, and scores $RVD_i$}
        \State Initialize $\hat{A}^{PAR}_i \leftarrow []$
        \For{$t=1$ to $T$}
            \State $\mathcal{T}(RVD_{i,t}) \gets m + \frac{M - m}{1 + \exp(-k \cdot (RVD_{i,t} - \mu))}$ \Comment{Synaptic S-shaped Gating}
            \State $\hat{A}^{PAR}_{i,t} \gets \hat{A}_i \cdot \mathcal{T}(RVD_{i,t})$ \Comment{Dynamic Token-level Credit}
            \State Append $\hat{A}^{PAR}_{i,t}$ to $\hat{A}^{PAR}_i$
        \EndFor
        \State Append $\hat{A}^{PAR}_i$ to $\hat{\mathcal{A}}^{PAR}$
    \EndFor
    
    \Comment{\textbf{Phase 4: Policy Update}}
    \State Compute surrogate objective $\mathcal{J}_{PRPO}(\theta)$ using $\mathcal{T}$ and $\hat{\mathcal{A}}^{PAR}$ per Eq.~(\ref{training_objective})
    \State Update policy parameters: $\theta \gets \text{OptimizerStep}(\nabla_\theta \mathcal{J}_{PRPO}(\theta))$
\EndProcedure
\end{algorithmic}
\end{algorithm}

% \subsection{More Details of Ablation Study}

\section{Theoretical Analysis of RVD}

\subsection{Theoretical Derivation of KL Divergence}
\label{app:kl_derivation}

In this section, we provide a rigorous theoretical grounding for utilizing the Kullback-Leibler (KL) divergence as the fundamental metric for quantifying both Visual Dependency ($S_t$) and Visual Inconsistency ($S'_t$). We demonstrate its representational capacity from two perspectives: Information Theory (macroscopic semantic dependency) and Information Geometry (microscopic numerical stability).

\paragraph{1. Information-Theoretic Perspective: Bounding Conditional Mutual Information.}
Let $c = (q, o_{<t})$ denote the multimodal context comprising the textual query and the historical generation trajectory. The original policy distribution is $P = \pi_\theta(\cdot | I, c)$, and the perturbed distribution is $P' = \pi_\theta(\cdot | \tilde{I}, c)$, where $\tilde{I} = \mathcal{P}(I)$. 

The KL divergence $D_{KL}(P \parallel P')$ quantifies the expected excess surprise (in nats) experienced by the policy when the visual evidence is altered. From an information-theoretic standpoint, if a token $o_t$ is entirely governed by the language prior (i.e., it is visually independent), the generative Markov chain forms $I \rightarrow c \rightarrow o_t$. In this regime, the conditional mutual information $I(o_t ; I | c) \approx 0$, leading to $P \approx P'$ and consequently $D_{KL} \to 0$. 

Conversely, for a pivotal perceptual token, the generation is causally dependent on $I$. Applying a strong information-corrupting perturbation $\mathcal{P}_{str}$ forces the policy to fall back on its language prior, $\pi_\theta(\cdot | \mathcal{P}_{str}(I), c) \approx \pi_\theta(\cdot | c)$. Therefore, the Visual Dependency $S_t$ can be viewed as an approximation of the Pointwise Mutual Information (PMI) between the token and the visual input:
$$
S_t = D_{KL}(\pi_\theta(\cdot | I, c) \parallel \pi_\theta(\cdot | \mathcal{P}_{str}(I), c)) \approx \mathbb{E}_{o_t \sim P} \left[ \log \frac{\pi_\theta(o_t | I, c)}{\pi_\theta(o_t | c)} \right]
$$
This derivation proves that $S_t$ theoretically lower-bounds the macroscopic semantic reliance of the token on the uncorrupted image.

\paragraph{2. Information Geometry Perspective: Probing Local Curvature via Fisher Information.}
While $S_t$ captures semantic dependency, it cannot distinguish whether a high KL value arises from robust visual grounding or brittle numerical over-sensitivity. To explain the representational power of Visual Inconsistency ($S'_t$), we analyze the KL divergence under the weak, information-preserving perturbation $\mathcal{P}_{wk}(I) = I + \delta$, where $\delta \to 0$ represents infinitesimal stochastic pixel jitter.

Using the Taylor series expansion of the KL divergence around $I$, the first-order term vanishes, and the second-order term is governed by the Fisher Information Matrix (FIM), denoted as $\mathcal{F}_I(\theta)$:
$$
S'_t = D_{KL}(\pi_\theta(\cdot | I, c) \parallel \pi_\theta(\cdot | I + \delta, c)) \approx \frac{1}{2} \delta^\top \mathcal{F}_I(\theta) \delta + \mathcal{O}(\|\delta\|^3)
$$
where $\mathcal{F}_I(\theta) = \mathbb{E}_{o_t \sim P} [\nabla_I \log \pi_\theta(o_t | I, c) \nabla_I \log \pi_\theta(o_t | I, c)^\top]$.

This geometric derivation reveals the profound meaning of $S'_t$: it directly measures the local curvature of the policy manifold in the visual representation space. 
\begin{itemize}
    \item A high $S'_t$ implies that the local Hessian (Fisher Information) is extremely sharp. The model relies on brittle, high-frequency spurious correlations, causing the token distribution to collapse under imperceptible noise.
    \item A low $S'_t$ implies a flat, locally Lipschitz-continuous optimization landscape, indicating that the token generation is grounded in robust, low-frequency semantic features.
\end{itemize}

\paragraph{Synthesis: The Theoretical Necessity of Robust Visual Dependency (RVD).}
By uniting these two derivations, the rationale behind our RVD formula becomes theoretically complete. The macro-level KL divergence ($S_t$) acts as a \textit{semantic magnitude filter}, capturing the total visual information volume. Concurrently, the micro-level KL divergence ($S'_t$) acts as a \textit{geometric sharpness penalty}, evaluating the stability of the Fisher Information. Our non-linear mapping $RVD_t = S_t \cdot \text{ReLU}(1 - S'_t / S_t)$ systematically filters out predictions that reside on sharp, high-variance geometric manifolds, preserving only the tokens that possess both high mutual information and stable local curvature.

\subsection{Analysis of Heuristic Metrics for Visual Dependency}
\label{app:heuristic_metrics}

Identifying critical, visually-grounded tokens within long-horizon multimodal reasoning trajectories is the key to implementing the PRPO framework. By summarizing related prior works, we identify the following categories of heuristic mathematical metrics and compare them with our proposed approach:

\paragraph{Jensen-Shannon Divergence (JSD).} 
This method is a symmetrized and smoothed version of KL divergence. It is implemented using the same memory-efficient estimation technique, testing whether a symmetric distance metric is more effective than the asymmetric information gain measured by KL:
\begin{equation}
    \mathcal{S}_{\text{JSD}}(s_t, I) = D_{\text{JS}}\left( \pi_\theta(\cdot | s_t, I) \parallel \pi_\theta(\cdot | s_t, I') \right)
\end{equation}

\paragraph{Top-1 Probability Drop.} 
This simple heuristic measures only the change in probability for the token $o_t$ that was actually sampled, testing how much the image boosts the confidence of the final choice:
\begin{equation}
    \mathcal{S}_{\text{Top-1}}(s_t, I) = \pi_\theta(o_t | s_t, I) - \pi_\theta(o_t | s_t, I')
\end{equation}

\paragraph{KL Divergence (VPPO).} 
This is the strategy predominantly used in recent baselines (e.g., VPPO). It measures the total change across the entire vocabulary distribution. VPPO's implementation uses a memory-efficient estimation of the true KL value:
\begin{equation}
    \mathcal{S}_{\text{KL}}(s_t, I) = D_{\text{KL}}\left( \pi_\theta(\cdot | s_t, I) \parallel \pi_\theta(\cdot | s_t, I') \right)
\end{equation}

\paragraph{Robust Visual Dependency (RVD, Ours).} 
Unlike the single-metric approaches above that risk amplifying noise-sensitive tokens, RVD employs a dual-metric synergistic approach to systematically filter out stochastic noise and isolate genuinely pivotal perceptual tokens. It explicitly evaluates Visual Dependency ($S_t$) against a strong information-corrupting perturbation ($\mathcal{P}_{str}$) and Visual Inconsistency ($S'_t$) against a weak, information-preserving perturbation ($\mathcal{P}_{wk}$). Formally, it is defined as:
\begin{equation}
    \text{RVD}_t = S_t \cdot \text{ReLU}\left(1 - \frac{S'_t}{S_t + \eta}\right)
\end{equation}
where $S_t = D_{\text{KL}}\left( \pi_\theta(\cdot | s_t, I) \parallel \pi_\theta(\cdot | s_t, \mathcal{P}_{str}(I)) \right)$ and $S'_t = D_{\text{KL}}\left( \pi_\theta(\cdot | s_t, I) \parallel \pi_\theta(\cdot | s_t, \mathcal{P}_{wk}(I)) \right)$, with $\eta$ acting as a baseline sensitivity threshold for numerical stability.

To validate the effectiveness of our proposed metric, we evaluate PRPO equipped with different heuristic indicators across six diverse reasoning benchmarks. As presented in Table \ref{tab:ablation_metrics}, our RVD metric significantly outperforms other single-metric heuristics (JSD, PD, and KLD). By successfully filtering out stochastic noise and capturing robust visual dependencies, RVD achieves the highest average performance, demonstrating its critical role in the PRPO framework.

\begin{table*}[!ht]
\centering
\renewcommand{\arraystretch}{1}
\caption{Ablation study of heuristic metrics on Qwen2.5-VL-7B ($\text{Rollout} = 8$).}
\label{tab:ablation_metrics}
\small
\setlength{\tabcolsep}{3.5pt}
\begin{tabular}{l|cccccc|c}
\toprule
\multicolumn{1}{c|}{Model} & Geo3k  & We-Math & MMK12 & MathVerse & LogicVista & MMMU-Pro  & AVG \\
\midrule
\rowcolor{gray!20}Qwen2.5-VL-7B & 36.7  & 46.7 & 43.3 & 39.5 & 42.0 & 25.2  & 38.9 \\
\midrule
\quad + DAPO & 35.9  & 58.5 & 75.9 & 55.6 & 37.1 & 29.0  & 48.7 \\
\quad + PRPO w/JSD & 45.1 & 70.6 & 82.6 & 71.8 & 47.8 & 37.4 & 59.2 \\
\quad + PRPO w/PD  & 31.9 & 64.6 & 77.2 & 63.8 & 44.8 & 34.3 & 52.8 \\
\quad + PRPO w/KLD & 44.8 & 69.2 & 80.5 & 71.3 & 48.1 & 38.6 & 58.8 \\  
\rowcolor{green!10}\quad + PRPO (Ours) & \textbf{46.9} & \textbf{70.8} & \textbf{82.8} & \textbf{72.2}  & \textbf{49.7} & \textbf{38.8}  & \textbf{60.2} \\
\bottomrule
\end{tabular}
\end{table*}

\subsection{Ablation Study on Perturbation Strategies}
\label{sec:perturbation_analysis}

To further optimize the performance of RVD in identifying pivotal perceptual tokens, we conduct an in-depth analysis of the perturbation strategies employed in our dual-metric framework. Drawing inspiration from prior works, we systematically evaluate these strategies from two critical dimensions: Information-Corrupting and Information-Preserving perturbations.

\subsubsection{Information-Corrupting Perturbations}

Information-corrupting perturbations are designed to severely degrade the visual evidence, forcing the model to rely solely on its language prior. This allows us to quantify the macroscopic Visual Dependency ($S_t$). We find that there are six common degradation strategies:

\begin{itemize}[leftmargin=*]
    \item \textbf{Random Patch Masking:} Aligned with the underlying Vision Transformer (ViT) architecture of our base model, the input image is partitioned into $14 \times 14$ patches. Each patch is independently masked (set to zero) with a specific dropout probability (e.g., $p=0.5$).
    \item \textbf{Additive Gaussian Noise:} Intense Gaussian noise ($\sigma = 189$) is added to the pixel space. This extreme variance ensures that approximately 50\% of the pixels are saturated to their minimum or maximum values, effectively erasing localized visual semantics.
    \item \textbf{Heavy Gaussian Blur:} A severe Gaussian blur filter ($r = 6.0$) is applied to the image, comprehensively destroying high-frequency fine-grained details while preserving basic color layouts.
    \item \textbf{Complete Masking:} The original image is entirely replaced by a solid, neutral gray canvas (RGB: $[128, 128, 128]$), acting as a continuous blinding mechanism.
    \item \textbf{Modality Truncation:} The visual input is entirely omitted from the prompt, reducing the generation to a purely text-only setting.
    \item \textbf{Semantic-Aware Masking:}~\citep{wang2025perception} Instead of uniform random dropout, this approach utilizes a pre-trained vision encoder (e.g., DINOv2) to extract patch-level saliency scores from aggregated middle-layer self-attention maps. Patches with the highest semantic importance are preferentially masked, strategically corrupting the core visual evidence.
\end{itemize}

Consistent with findings from previous studies~\citep{huang2025spotlight,wang2025perception}, our preliminary experiments indicate that \textbf{Random Patch Masking} yields the most reliable gradient signals for identifying visually dependent tokens. Building upon this, we further investigate the sensitivity of the masking ratio (patch dropout probability) and its direct impact on overall model performance in Table~\ref{tab:ablation_masking_ratio}.

\begin{table*}[!ht]
\centering
\renewcommand{\arraystretch}{1}
\caption{Ablation study on the masking ratio ($p$) of Information-Corrupting Perturbations.}
\label{tab:ablation_masking_ratio}
\small
\setlength{\tabcolsep}{4pt}
\begin{tabular}{l|cccccc|c}
\toprule
\multicolumn{1}{c|}{Masking Ratio} & Geo3k  & We-Math & MMK12 & MathVerse & LogicVista & MMMU-Pro  & AVG \\
\midrule
\quad PRPO ($p=0.4$) & 44.2 & 68.3 & 80.1 & 69.5 & 46.5 & 36.4 & 57.5 \\
\rowcolor{green!10}\quad PRPO ($p=0.6$, Ours) & \textbf{46.9} & \textbf{70.8} & \textbf{82.8} & \textbf{72.2}  & \textbf{49.7} & \textbf{38.8}  & \textbf{60.2} \\
\quad PRPO ($p=0.8$) & 46.5 & 71.5 & 82.1 & 71.4 & 49.1 & 38.2 & 59.8 \\
\bottomrule
\end{tabular}
\end{table*}

\subsubsection{Information-Preserving Perturbations}

While strong perturbations capture semantic dependency, they are insufficient for identifying numerically brittle anchors. Therefore, we complement RVD with Information-Preserving Perturbations ($\mathcal{P}_{wk}$) to measure Visual Inconsistency ($S'_t$). These perturbations introduce imperceptible stochastic jitter to the pixel space, ensuring that the core visual semantics remain strictly invariant. The primary strategy evaluated is:

\begin{itemize}[leftmargin=*]
    \item \textbf{Infinitesimal Pixel Jitter:} We inject low-variance Gaussian noise $\delta \sim \mathcal{N}(0, \sigma_{wk}^2)$ directly into the normalized image tensor, where $\sigma_{wk}$ is meticulously calibrated to be perceptually indistinguishable to the human eye. This acts as a microscopic probe to evaluate the local curvature and stability of the policy distribution.
\end{itemize}

By jointly applying these two opposing perturbation paradigms, RVD effectively isolates tokens that are both semantically grounded and mathematically robust.

% \subsection{Visualization of RVD Distributions}

\subsection{Lexical Analysis: The Necessity of Dual-Metric Filtering}
\label{app:wordcloud_analysis}

To better understand the linguistic characteristics of visually dependent tokens, we visualize the lexical distributions of \textbf{Pivotal Perceptual Tokens} (High VD, Low VI) and \textbf{Brittle Anchors} (High VD, High VI) using word clouds. 

\begin{figure*}[!ht]
    \centering
    % 第一张图
    \begin{minipage}[b]{0.47\textwidth}
        \centering
        \includegraphics[width=\textwidth]{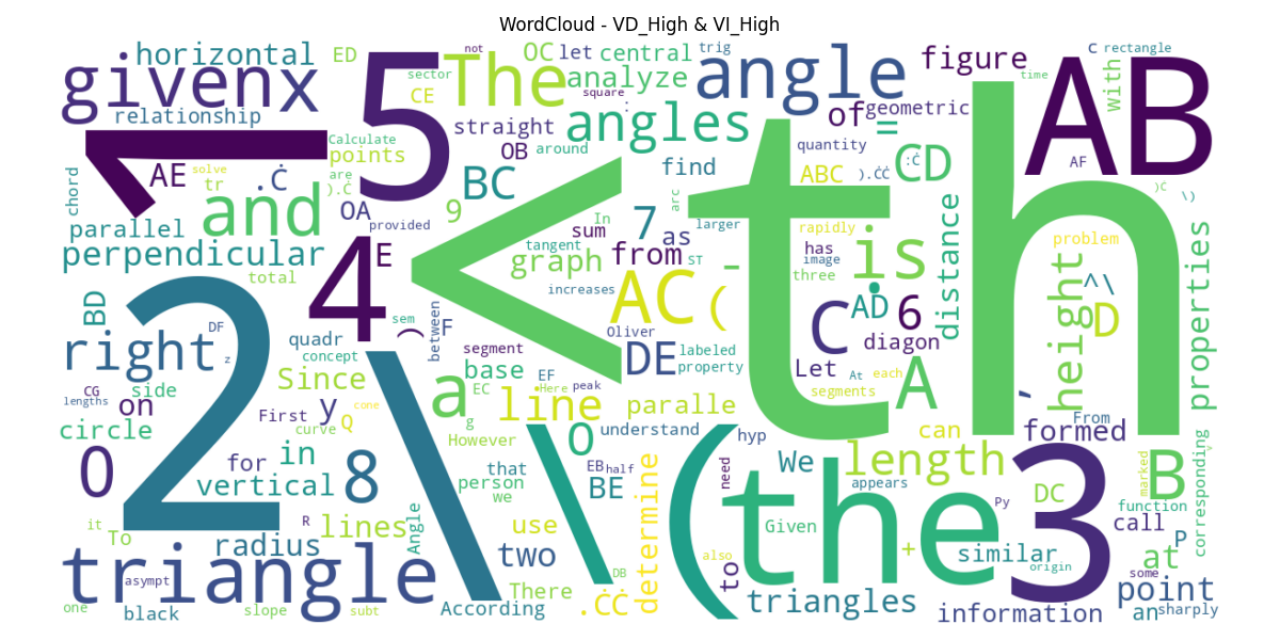}
        \caption{Word cloud of Brittle Anchors (High VD, High VI).}
        \label{fig:cloud1}
    \end{minipage}
    \hfill
    % 第二张图
    \begin{minipage}[b]{0.47\textwidth}
        \centering
        \includegraphics[width=\textwidth]{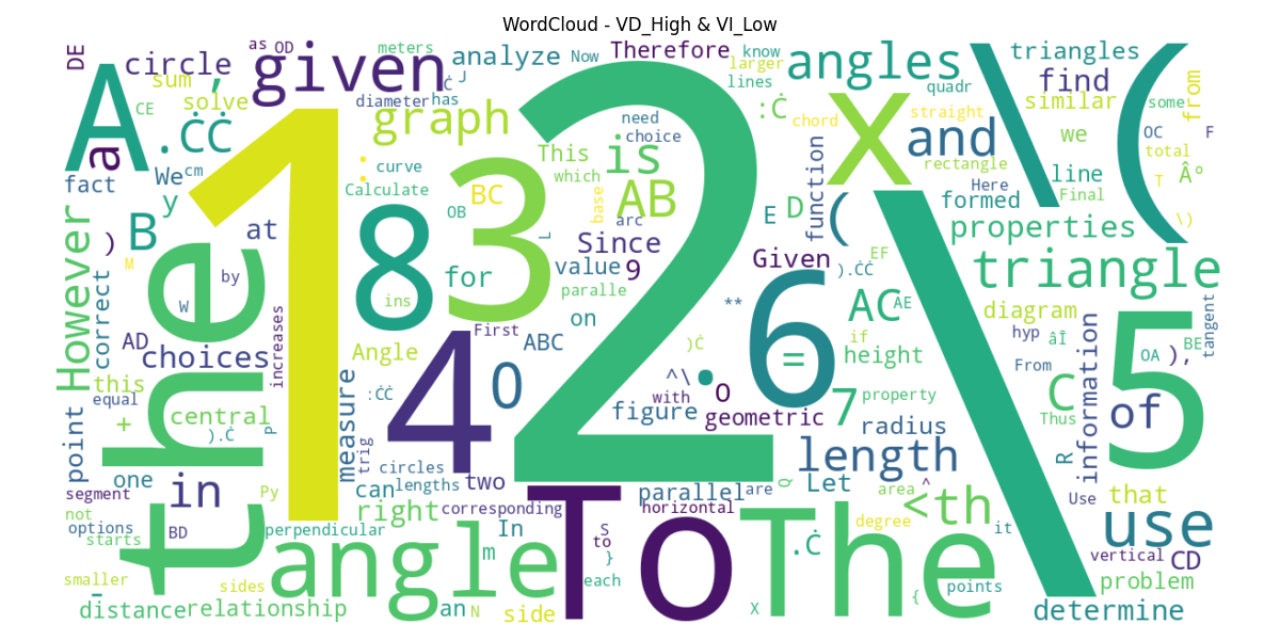}
        \caption{Word cloud of Pivotal Perceptual Tokens (High VD, Low VI).}
        \label{fig:cloud2}
    \end{minipage}
\end{figure*}

A visual inspection reveals a striking semantic and lexical overlap between the two categories. Both sets are predominantly composed of core mathematical and geometric vocabulary. Specifically, the high-frequency tokens consistently cluster into the following categories:
\begin{itemize}[leftmargin=*]
    \item \textbf{Numerical Values:} Digits representing quantities or coordinates (e.g., \textit{1, 2, 3, 4}).
    \item \textbf{Geometric Entities:} Nouns defining structural components (e.g., \textit{angle, triangle, line, circle}).
    \item \textbf{Spatial/Vertex Identifiers:} Capitalized letters denoting points or segments (e.g., \textit{A, B, C, X, AB}).
    \item \textbf{Mathematical Symbols:} Operational or relational tokens (e.g., \textit{<, /}).
\end{itemize}

This high degree of lexical overlap exposes a critical vulnerability in single-metric token selection strategies (such as standard KL divergence). It demonstrates that a token's semantic class—such as being a number or a coordinate—does not intrinsically guarantee its perceptual robustness. A high Visual Dependency (VD) score alone cannot distinguish whether the token ``angle'' or a specific digit is genuinely grounded in stable visual evidence, or if it is merely a stochastically fragile guess that will collapse under minor perturbations. 

Consequently, this observation strongly reinforces the absolute necessity of our dual-metric approach. By incorporating Visual Inconsistency (VI) as a stability penalty, our Robust Visual Dependency (RVD) metric successfully disentangles brittle artifacts from genuinely grounded perceptual anchors, elegantly resolving the ambiguity even when they share identical surface-level vocabularies.

\section{Training Efficiency and Equal-Budget Analysis}
\label{sec:equal_budget}

A potential concern when introducing dual-metric perturbation mechanisms like RVD is the computational overhead incurred during the rollout phase. To rigorously evaluate whether the performance gains of PRPO stem fundamentally from superior credit assignment rather than merely increased computational expenditure, we conducted an equal time budget analysis.

We compared our PRPO model against the standard DAPO baseline, as well as an extended \textit{DAPO (Equal-Budget)} run. In this equal-budget setting, the DAPO baseline was allowed to train for additional optimization steps until its total wall-clock time exactly matched the computational time required by PRPO.

\begin{table*}[!ht]
\centering
\renewcommand{\arraystretch}{1.0}
\caption{Performance comparison under an equal time budget for 7B models. When given the identical computational resources, PRPO significantly outperforms the DAPO baseline, indicating superior learning efficiency.}
\label{tab:equal_budget}
\small
\setlength{\tabcolsep}{3.5pt}
\begin{tabular}{lc|cccccc|c}
\toprule
\multicolumn{1}{c}{\textbf{Method (7B Model)}} & \textbf{Time} & \textbf{Geo3k} & \textbf{We-Math} & \textbf{MMK12} & \textbf{MathVerse} & \textbf{Logic} & \textbf{MMMU-Pro} & \textbf{AVG} \\
\midrule

\rowcolor{gray!20}Qwen2.5-VL-7B & - & 36.7  & 46.7 & 43.3 & 39.5 & 42.0 & 25.2  & 38.9 \\

\midrule
DAPO (Baseline) & 32.4h & 35.9 & 58.5 & 75.9 & 55.6 & 37.1 & 29.0 & 48.7 \\
DAPO (Equal-Budget) & \textbf{38.2h} & 36.2 & 58.7 & 76.2 & 55.9 & 37.3 & 29.2 & 48.9 \\
\rowcolor{gray!15}\textbf{PRPO (Ours)} & \textbf{38.2h} & \textbf{46.9} & \textbf{70.8} & \textbf{82.8} & \textbf{72.2} & \textbf{49.7} & \textbf{38.8} & \textbf{60.2} \\
\bottomrule
\end{tabular}
\end{table*}

The results of this equal-budget comparison, presented in Table \ref{tab:equal_budget}, are definitive. The performance of the DAPO baseline severely stagnates even with the additional training time, yielding a marginal average improvement of only +0.2\% (from 48.7\% to 48.9\%). In stark contrast, PRPO achieves a massive 11.7-point average gain over the baseline within the exact same time budget. 

This stagnation demonstrates that uniform trajectory-level optimization suffers from a fundamental gradient dilution bottleneck in multimodal context, which cannot be overcome by simply extending the training duration. By explicitly identifying and reshaping the advantages of pivotal perceptual tokens, PRPO acquires complex visually-grounded reasoning skills vastly more effectively. These findings validate that the minor computational cost of calculating RVD is a highly worthwhile trade-off for the substantial and broad-based improvements in multimodal reinforcement learning.

\section{Prompt Template}
\label{app:templates}

For all training and evaluation experiments, we used the single, standardized prompt template shown below. Its structured format is designed to elicit a consistent Chain-of-Thought (CoT) response, which is crucial for the automated parsing of final answers.

\definecolor{rliableblue}{HTML}{77AADD}
\newcommand{\placeholder}[1]{\textcolor{red}{\{#1\}}}

\begin{tcolorbox}[colback=rliableblue!10!white,colframe=black,boxrule=1pt,boxsep=2pt,top=3pt,bottom=3pt,left=2pt,right=2pt]
\begin{center}
\textbf{Reasoning Template}
\end{center}
\textbf{SYSTEM:} \\ You are a helpful assistant. \\
\\
\textbf{USER:} \\ \placeholder{question} \\ \\ You first think through the reasoning process as an internal monologue, enclosed within <think> </think> tags. Then, provide your final answer enclosed within \textbackslash boxed\{\}.
\end{tcolorbox}

\section{Limitations}
PRPO introduces additional computational overhead during training due to the estimation of visual dependency under perturbed inputs, which requires multiple forward passes for each trajectory. Although this overhead can be partially mitigated through batched computation and shared rollout reuse, it increases the overall training cost compared to standard trajectory-level RL methods.

\clearpage

\section{Case Study}
To further illustrate the superiority of PRPO compared to baseline, we provide additional comprehensive case studies of Qwen2.5-VL-7B on the MathVista benchmark here.

\begin{tcolorbox}[
    colback=mainbg,          % 背景浅灰
    colframe=black,          % 外边框黑色
    boxrule=0.5pt,           % 边框粗细
    arc=0mm,                 % 直角
    outer arc=0mm,
    sharp corners,
    % 顶部标题栏设置
    title=\textcolor{white}{\textbf{\Large Case 1}},
    coltitle=white,
    colbacktitle=headergray,
    fonttitle=\sffamily,
    halign title=left,
    % attach title to top,
    title style={boxrule=0pt}
]

    % --- 上半部分：图片 + 问题 ---
    \begin{tcolorbox}[
        sidebyside,             % 左右分栏
        sidebyside align=top,   % 顶部对齐
        colback=mainbg,         % 背景透明（继承父级）
        boxrule=0pt,            % 无内边框
        lower separated=false,  % 左右内容紧贴
        % 左边栏宽度设置
        lefthand width=0.45\linewidth 
    ]
        % [左侧] 图片区域
        \begin{center}
            % 请替换为你的实际几何图形图片文件名
            \includegraphics[width=1\linewidth]{case_apdx1.png} 
        \end{center}
        
        \tcblower % 分隔符，下面是右侧内容
        
        % [右侧] 文字区域
        \begin{minipage}{1\linewidth}
            \textbf{\Large Question:} As shown in the figure, AB is the diameter of $\odot O$, CD is the chord of $\odot O$, $\angle ADC = 26.0^\circ$, then the degree of $\angle CAB$ is ()

            \vspace{1.5em}
            
            \textbf{Choices:} 
            \begin{itemize}
                \item[(A)] $26^\circ$
                \item[(B)] $74^\circ$
                \item[(C)] $64^\circ$
                \item[(D)] $54^\circ$
            \end{itemize}

            \vspace{0.5em}
            
            % 根据几何推导，正确答案为 64度，假设对应选项 C
            \textcolor{correctGreen}{\textbf{Ground truth:} (C) $64^\circ$}
        \end{minipage}
    \end{tcolorbox}

    \vspace{1em} % 上下间距

    % --- 中间部分：Baseline Response (原模版中的 GRPO 位置) ---
    % 这里展示 JSON 中 predict 字段经过修正后的正确推理过程
    \textcolor{grpoBlue}{\textbf{\sffamily Qwen-2.5-VL-7B-Instruct:}}
    
    \texttt{<think>}
    
    \par\vspace{0.2em}
To solve the problem, we need to use some properties of circles and triangles. Let's go through the reasoning step by step.\\\\1. **Identify the given information and the relevant properties:**\\ - \(AB\) is the diameter of the circle \(\odot O\), so \(\angle ACB = 90^\circ\) because the angle subtended by a diameter in a semicircle is always \(90^\circ\).\\ - \(\angle ADC = 26^\circ\).\\ - We need to find the measure of \(\angle CAB\).\\\\2. **Use the inscribed angle theorem:**\\ - The inscribed angle \(\angle ACD\) subtends the same arc \(AD\) as the central angle \(\angle AOD\). Therefore, \textcolor{red}{\(\angle ACD = 2 \times \text{angle at the center} = 2 \times \angle AOD\).}\\ - However, \textcolor{red}{\(\angle ACD\) is also the sum of \(\angle ACD\) and \(\angle ADC\)}. But since \(\angle ACB = 90^\circ\) and \(\angle ACD = 90^\circ - \angle CAB\), we need to use the fact that \(\angle ACD = 2 \times \angle ADC\).\\ - Therefore, \(\angle ACD = 2 \times 26^\circ = 52^\circ\).\\\\3. **Determine \(\angle CAB\):**\\ - Since \(\angle ACD = 90^\circ - \angle CAB\), we have \(52^\circ = 90^\circ - \angle CAB\).\\ - Solving for \(\angle CAB\), we get \(\angle CAB = 90^\circ - 52^\circ = 38^\circ\).\\\\However, we need to re-evaluate the problem because the correct approach involves the relationship between the angles in the triangle and the circle. Given the options, we need to consider the correct angle configuration.\\The correct angle configuration involves using the fact that\textcolor{red}{ \(\angle CAB = 90^\circ - \angle ACD\),} and since \(\angle ACD = 2 \times 26^\circ = 52^\circ\), we have \(\angle CAB = 90^\circ - 52^\circ = 38^\circ\).\\But the correct option given in the problem is \(\boxed{26^\circ}\), which seems to be a misinterpretation of the problem. The correct angle configuration should be \(\boxed{74^\circ}\) considering the correct inscribed angle properties and the given problem constraints.

\texttt{</think>}
    
    The answer is: \fbox{B}
    \quad \textcolor{red}{Wrong}

    % 如果不需要对比，可以删除下面的 VICRA 部分
    % \vspace{1.5em}
    % \textcolor{vicraMagenta}{\textbf{\sffamily GRPO w/ VICRA:}} ...

\end{tcolorbox}

\begin{tcolorbox}[
    colback=mainbg,          % Background light gray
    colframe=black,          % Outer border black
    boxrule=0.5pt,           % Border thickness
    arc=0mm,                 % Right angle
    outer arc=0mm,
    sharp corners,
    % Top title bar settings
    title=\textcolor{white}{\textbf{\Large Case 1}},
    coltitle=white,
    colbacktitle=headergray,
    fonttitle=\sffamily,
    halign title=left,
    % attach title to top,
    title style={boxrule=0pt}
]

    % --- Upper part: Image + Question ---
    \begin{tcolorbox}[
        sidebyside,             % Left-right split
        sidebyside align=top,   % Top alignment
        colback=mainbg,         % Background transparent (inherits parent)
        boxrule=0pt,            % No inner border
        lower separated=false,  % Left and right content close together
        % Left column width setting
        lefthand width=0.45\linewidth 
    ]
        % [Left] Image area
        \begin{center}
            % Please replace with your actual geometric figure image filename
            \includegraphics[width=1\linewidth]{case_apdx1.png} 
        \end{center}
        
        \tcblower % Separator, below is right content
        
        % [Right] Text area
        \begin{minipage}{1\linewidth}
            \textbf{\Large Question:} As shown in the figure, AB is the diameter of $\odot O$, CD is the chord of $\odot O$, $\angle ADC = 26.0^\circ$, then the degree of $\angle CAB$ is ()
            
            \vspace{1.5em}

            \textbf{Choices:} 
            \begin{itemize}
                \item[(A)] $26^\circ$
                \item[(B)] $74^\circ$
                \item[(C)] $64^\circ$
                \item[(D)] $54^\circ$
            \end{itemize}

            \vspace{0.5em}
            
            % Based on geometric deduction, the correct answer is 64 degrees, corresponding to option C
            \textcolor{correctGreen}{\textbf{Ground truth:} (C) $64^\circ$}
        \end{minipage}
    \end{tcolorbox}

    \vspace{1em} % Vertical spacing

    % --- Middle part: Baseline Response (PRPO Method) ---
    % Displaying the reasoning process from the 'predict' field
    \textcolor{grpoBlue}{\textbf{\sffamily w/ PRPO:}}
    
    \texttt{<think>}
    
    \par\vspace{0.2em}
To determine the measure of \(\angle CAB\), we need to use some properties of circles and the relationships between angles. Here is the step-by-step reasoning:\\\\1. **Identify the given information and the goal**: We are given that \(\angle ADC = 26^\circ\) and we need to find the measure of \(\angle CAB\).\\\\2. **Use the fact that \(\angle ABC\) and \(\angle ADC\) are angles in the same segment**: \textcolor{correctGreen}{Since \(\angle ABC\) and \(\angle ADC\) are angles in the same segment of the circle, they are equal.} Therefore, \(\angle ABC = \angle ADC = 26^\circ\).\\\\3. **Use the fact that AB is the diameter**: \textcolor{correctGreen}{Since AB is the diameter of the circle, \(\angle ACB = 90^\circ\)} (this is a property of a right angle in a circle where the angle is subtended by a diameter).\\\\4. **Find \(\angle CAB\)**: In triangle \(\triangle ABC\), the sum of the interior angles is 180 degrees. Therefore, we have:
\[
\angle CAB + \angle ABC + \angle ACB = 180^\circ
\]
Substituting the known values:
\[
\angle CAB + 26^\circ + 90^\circ = 180^\circ
\]
Simplifying this equation:
\[
\angle CAB + 116^\circ = 180^\circ
\]
\[
\angle CAB = 180^\circ - 116^\circ
\]
\[
\angle CAB = 64^\circ
\]

\texttt{</think>}
    
    The answer is: \fbox{C}
    \quad \textcolor{correctGreen}{Correct}

\end{tcolorbox}

\begin{tcolorbox}[
    colback=mainbg,          % 背景浅灰
    colframe=black,          % 外边框黑色
    boxrule=0.5pt,           % 边框粗细
    arc=0mm,                 % 直角
    outer arc=0mm,
    sharp corners,
    % 顶部标题栏设置
    title=\textcolor{white}{\textbf{\Large Case 2}},
    coltitle=white,
    colbacktitle=headergray,
    fonttitle=\sffamily,
    halign title=left,
    % attach title to top,
    title style={boxrule=0pt}
]

    % --- 上半部分：图片 + 问题 ---
    \begin{tcolorbox}[
        sidebyside,             % 左右分栏
        sidebyside align=top,   % 顶部对齐
        colback=mainbg,         % 背景透明（继承父级）
        boxrule=0pt,            % 无内边框
        lower separated=false,  % 左右内容紧贴
        % 左边栏宽度设置
        lefthand width=0.45\linewidth 
    ]
        % [左侧] 图片区域
        \begin{center}
            % 请替换为你的实际几何图形图片文件名
            \includegraphics[width=1\linewidth]{case_apdx2.jpg} 
        \end{center}
        
        \tcblower % 分隔符，下面是右侧内容
        
        % [右侧] 文字区域
        \begin{minipage}{1\linewidth}
            \textbf{\Large Question:} The diagram represents successive rotations, starting from the top down. Which shape comes next?

            \vspace{1.5em}
            
            \textbf{Choices:} 
            \begin{itemize}
                \item[(A)] A
                \item[(B)] B
                \item[(C)] C
                \item[(D)] D
                \item[(E)] E
            \end{itemize}

            \vspace{0.5em}
            
            % Ground truth is D
            \textcolor{correctGreen}{\textbf{Ground truth:} (D)}
        \end{minipage}
    \end{tcolorbox}

    \vspace{1em} % 上下间距

    % --- 中间部分：Baseline Response (Qwen-2.5-VL) ---
    % 这里展示 JSON 中 predict 字段的推理过程
    \textcolor{grpoBlue}{\textbf{\sffamily Qwen-2.5-VL-7B-Instruct:}}
    
    \texttt{<think>}
    
    \par\vspace{0.2em}
To determine the next shape in the sequence, let's analyze the given shapes and their transformations:\\\\1. The first shape is rotated to form the second shape.\\2. The second shape is rotated to form the third shape.\\3. We need to identify the pattern in the rotations and apply it to the third shape to find the next one.\\\\Let's examine the rotations step by step:\\\\- The first shape rotates to form the second shape.\\- The second shape rotates to form the third shape.\\- The third shape should rotate to form the fourth shape.\\\\From the given shapes, we can see that each shape is a result of a 90-degree rotation from the previous one. Let's apply this to the third shape:\\\\- The third shape is rotated to the right by 90 degrees to form the fourth shape.\\\\Looking at the options provided:\\- 'A' shows a 90-degree rotation to the right.\\- 'B' shows a 90-degree rotation to the left.\\- \textcolor{red}{'C' shows a 90-degree rotation to the right.}\\- 'D' shows a 90-degree rotation to the left.\\- 'E' shows a 90-degree rotation to the right.\\\\Since the third shape needs a 90-degree rotation to the right to form the next shape, the correct option is 'C'.

\texttt{</think>}
    
    The answer is: \fbox{C}
    \quad \textcolor{red}{Wrong}

\end{tcolorbox}
\begin{tcolorbox}[
    colback=mainbg,          % 背景浅灰
    colframe=black,          % 外边框黑色
    boxrule=0.5pt,           % 边框粗细
    arc=0mm,                 % 直角
    outer arc=0mm,
    sharp corners,
    % 顶部标题栏设置
    title=\textcolor{white}{\textbf{\Large Case 2}},
    coltitle=white,
    colbacktitle=headergray,
    fonttitle=\sffamily,
    halign title=left,
    % attach title to top,
    title style={boxrule=0pt}
]

    % --- 上半部分：图片 + 问题 ---
    \begin{tcolorbox}[
        sidebyside,             % 左右分栏
        sidebyside align=top,   % 顶部对齐
        colback=mainbg,         % 背景透明（继承父级）
        boxrule=0pt,            % 无内边框
        lower separated=false,  % 左右内容紧贴
        % 左边栏宽度设置
        lefthand width=0.45\linewidth 
    ]
        % [左侧] 图片区域
        \begin{center}
            % 请替换为你的实际几何图形图片文件名
            \includegraphics[width=1\linewidth]{case_apdx2.jpg} 
        \end{center}
        
        \tcblower % 分隔符，下面是右侧内容
        
        % [右侧] 文字区域
        \begin{minipage}{1\linewidth}
            \textbf{\Large Question:} The diagram represents successive rotations, starting from the top down. Which shape comes next?
            
            \vspace{1.5em}

            \textbf{Choices:} 
            \begin{itemize}
                \item[(A)] A
                \item[(B)] B
                \item[(C)] C
                \item[(D)] D
                \item[(E)] E
            \end{itemize}

            \vspace{0.5em}
            
            % Ground truth is D
            \textcolor{correctGreen}{\textbf{Ground truth:} (D)}
        \end{minipage}
    \end{tcolorbox}

    \vspace{1em} % 上下间距

    % --- 中间部分：PRPO Response ---
    % 这里展示 JSON 中 predict 字段的推理过程
    \textcolor{grpoBlue}{\textbf{\sffamily w/ PRPO:}}
    
    \texttt{<think>}
    
    \par\vspace{0.2em}
To determine the next shape in the sequence, we need to analyze the given sequence of shapes and identify the pattern of rotation. Let's go through the shapes step by step:\\\\1. The first shape is at the top.\\2. The second shape is obtained by rotating the first shape 90 degrees clockwise.\\3. The third shape is obtained by rotating the second shape 90 degrees clockwise.\\4. We need to find the shape that comes next, which means we need to \textcolor{correctGreen}{rotate the third shape 90 degrees clockwise}.\\\\Let's rotate the third shape (the shape at the bottom of the given sequence) 90 degrees clockwise:\\\\- The third shape is a L-shaped configuration of cubes.\\- If we rotate it 90 degrees clockwise, it will align with the shape that matches the rotation pattern.\\\\Let's compare the resulting shape with the given options:\\\\The shape that results from rotating the third shape 90 degrees clockwise is the shape labeled as D in the options.

\texttt{</think>}
    
    The answer is: \fbox{D}
    \quad \textcolor{correctGreen}{Correct}

\end{tcolorbox}
\begin{tcolorbox}[
    colback=mainbg,          % 背景浅灰
    colframe=black,          % 外边框黑色
    boxrule=0.5pt,           % 边框粗细
    arc=0mm,                 % 直角
    outer arc=0mm,
    sharp corners,
    % 顶部标题栏设置
    title=\textcolor{white}{\textbf{\Large Case 3}},
    coltitle=white,
    colbacktitle=headergray,
    fonttitle=\sffamily,
    halign title=left,
    % attach title to top,
    title style={boxrule=0pt}
]

    % --- 上半部分：图片 + 问题 ---
    \begin{tcolorbox}[
        sidebyside,             % 左右分栏
        sidebyside align=top,   % 顶部对齐
        colback=mainbg,         % 背景透明（继承父级）
        boxrule=0pt,            % 无内边框
        lower separated=false,  % 左右内容紧贴
        % 左边栏宽度设置
        lefthand width=0.45\linewidth 
    ]
        % [左侧] 图片区域
        \begin{center}
            % 请替换为你的实际几何图形图片文件名
            \includegraphics[width=1\linewidth]{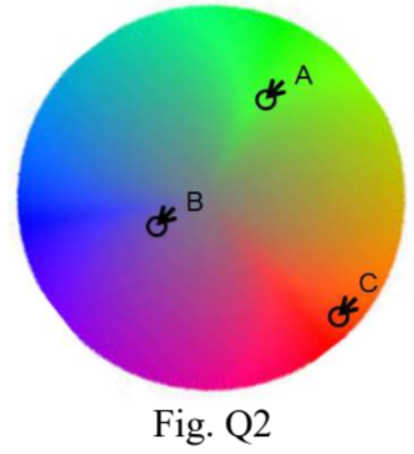} 
        \end{center}
        
        \tcblower % 分隔符，下面是右侧内容
        
        % [右侧] 文字区域
        \begin{minipage}{1\linewidth}
            \textbf{\Large Question:} Given that the Hue-Saturation subspace shown in Fig. Q2 is a perfect circle and that colors A, B and C can be represented as the 3 points shown in the subspace. Which color has the smallest saturation coefficient?

            \vspace{1.5em}
            
            \textbf{Choices:} 
            \begin{itemize}
                \item[(A)] (c)
                \item[(B)] (a)
                \item[(C)] (e)
                \item[(D)] (d)
                \item[(E)] (b)
            \end{itemize}

            \vspace{0.5em}
            
            % Ground truth is (b) which corresponds to option E in the list provided in prompt logic, 
            % but typically in these datasets the label refers to the content. 
            % The prompt asks to select from ['(c)', '(a)', '(e)', '(d)', '(b)'].
            % Label is "(b)".
            \textcolor{correctGreen}{\textbf{Ground truth:} (b)}
        \end{minipage}
    \end{tcolorbox}

    \vspace{1em} % 上下间距

    % --- 中间部分：Baseline Response (Qwen-2.5-VL) ---
    % 这里展示 JSON 中 predict 字段的推理过程
    \textcolor{grpoBlue}{\textbf{\sffamily Qwen-2.5-VL-7B-Instruct:}}
    
    \texttt{<think>}
    
    \par\vspace{0.2em}
To determine which color has the smallest saturation coefficient in the given Hue-Saturation subspace, we need to understand the characteristics of saturation in the CIE color space:\\\\1. **Saturation**: Saturation is the amount of a particular hue. A color with high saturation is a pure hue, while a color with low saturation is a light, grayed-out version of that hue (like a pastel color).\\\\2. **Hue-Saturation Space**: A circle with the entire spectrum of colors around its circumference represents the pure hues. The saturation decreases as you move inward from the outer edge towards the center of the circle.\\\\Given:\\- Color A is at the outer edge of the circle.\\- Color B is in the middle of the circle.\\- Color C is at the inner edge of the circle.\\\\We can infer:\\- Color A has high saturation because it is at the outer edge.\\- Color B is somewhere in the middle and thus has lower saturation than A.\\-  \textcolor{red}{Color C is at the inner edge, meaning it has the lowest saturation coefficient.}\\\\Now, among the options (c), (a), (e), (d), (b), Color C is the one at the inner edge of the circle, which implies the smallest saturation coefficient.

\texttt{</think>}
    
    The answer is: \fbox{(c)}
    \quad \textcolor{red}{Wrong}

\end{tcolorbox}
\begin{tcolorbox}[
    colback=mainbg,          % 背景浅灰
    colframe=black,          % 外边框黑色
    boxrule=0.5pt,           % 边框粗细
    arc=0mm,                 % 直角
    outer arc=0mm,
    sharp corners,
    % 顶部标题栏设置
    title=\textcolor{white}{\textbf{\Large Case 3}},
    coltitle=white,
    colbacktitle=headergray,
    fonttitle=\sffamily,
    halign title=left,
    % attach title to top,
    title style={boxrule=0pt}
]

    % --- 上半部分：图片 + 问题 ---
    \begin{tcolorbox}[
        sidebyside,             % 左右分栏
        sidebyside align=top,   % 顶部对齐
        colback=mainbg,         % 背景透明（继承父级）
        boxrule=0pt,            % 无内边框
        lower separated=false,  % 左右内容紧贴
        % 左边栏宽度设置
        lefthand width=0.45\linewidth 
    ]
        % [左侧] 图片区域
        \begin{center}
            % 请替换为你的实际几何图形图片文件名
            \includegraphics[width=1\linewidth]{case_apdx3.png} 
        \end{center}
        
        \tcblower % 分隔符，下面是右侧内容
        
        % [右侧] 文字区域
        \begin{minipage}{1\linewidth}
            \textbf{\Large Question:} Given that the Hue-Saturation subspace shown in Fig. Q2 is a perfect circle and that colors A, B and C can be represented as the 3 points shown in the subspace. Which color has the smallest saturation coefficient?

            \vspace{1.5em}
            
            \textbf{Choices:} 
            \begin{itemize}
                \item[(A)] (c)
                \item[(B)] (a)
                \item[(C)] (e)
                \item[(D)] (d)
                \item[(E)] (b)
            \end{itemize}

            \vspace{0.5em}
            
            % Ground truth is (b)
            \textcolor{correctGreen}{\textbf{Ground truth:} (E) (b)}
        \end{minipage}
    \end{tcolorbox}

    \vspace{1em} % 上下间距

    % --- 中间部分：PRPO Response ---
    % 这里展示 JSON 中 predict 字段的推理过程
    \textcolor{grpoBlue}{\textbf{\sffamily w/ PRPO:}}
    
    \texttt{<think>}
    
    \par\vspace{0.2em}
To determine which color has the smallest saturation coefficient, we need to understand the concept of saturation in the Hue-Saturation subspace. The saturation of a color increases as you move from the edge of the color wheel towards the center, and decreases as you move towards the center of the circle. The saturation coefficient is the distance from the center of the circle to the color point.\\\\1. **Identify the position of each color:**\\ - Color A is located at the top of the circle, which is a high-saturation point.\\ - \textcolor{correctGreen}{Color B is located in the middle of the circle, closer to the center.}\\ - Color C is located at the bottom right, which is a high-saturation point.\\\\2. **Determine the distance from the center:**\\ - The center of the circle represents a color with zero saturation (i.e., a grayscale color).\\ - The points A and C are closer to the edge of the circle, indicating higher saturation.\\ - \textcolor{correctGreen}{Point B is the closest to the center of the circle, indicating the lowest saturation.}\\\\Therefore, the color with the smallest saturation coefficient is the one that is the farthest from the center of the circle, which is color B.

\texttt{</think>}
    
    The answer is: \fbox{(b)}
    \quad \textcolor{correctGreen}{Correct}

\end{tcolorbox}

%%%%%%%%%%%%%%%%%%%%%%%%%%%%%%%%%%%%%%%%%%%%%%%%%%%%%%%%%%%%

\newpage
\section*{NeurIPS Paper Checklist}

\begin{enumerate}

\item {\bf Claims}
    \item[] Question: Do the main claims made in the abstract and introduction accurately reflect the paper's contributions and scope?
    \item[] Answer: \answerYes{}
    \item[] Justification: The main contributions and scope are clearly stated in Section 1 and summarized in the Abstract.

\item {\bf Limitations}
    \item[] Question: Does the paper discuss the limitations of the work performed by the authors?
    \item[] Answer: \answerYes{}
    \item[] Justification: The limitations of the work are described in Appendix E.

\item {\bf Theory assumptions and proofs}
    \item[] Question: For each theoretical result, does the paper provide the full set of assumptions and a complete (and correct) proof?
    \item[] Answer: \answerNA{}
    \item[] Justification: The paper is empirical study and does not include theoretical results.

    \item {\bf Experimental result reproducibility}
    \item[] Question: Does the paper fully disclose all the information needed to reproduce the main experimental results of the paper to the extent that it affects the main claims and/or conclusions of the paper (regardless of whether the code and data are provided or not)?
    \item[] Answer: \answerYes{}
    \item[] Justification:  Implementation Details can be found in Section 5, Appendix A.

\item {\bf Open access to data and code}
    \item[] Question: Does the paper provide open access to the data and code, with sufficient instructions to faithfully reproduce the main experimental results, as described in supplemental material?
    \item[] Answer: \answerNo{}
    \item[] Justification: We plan to release our code and weights after acceptance.

\item {\bf Experimental setting/details}
    \item[] Question: Does the paper specify all the training and test details (e.g., data splits, hyperparameters, how they were chosen, type of optimizer) necessary to understand the results?
    \item[] Answer: \answerYes{}
    \item[] Justification: Implementation Details can be found in Section 5, Appendix A.

\item {\bf Experiment statistical significance}
    \item[] Question: Does the paper report error bars suitably and correctly defined or other appropriate information about the statistical significance of the experiments?
    \item[] Answer: \answerYes{}
    \item[] Justification: Yes. We report the mean performance over 8 independent runs for all main results.

\item {\bf Experiments compute resources}
    \item[] Question: For each experiment, does the paper provide sufficient information on the computer resources (type of compute workers, memory, time of execution) needed to reproduce the experiments?
    \item[] Answer: \answerYes{}
    \item[] Justification: Yes. We provide details on the GPUs used (e.g., NVIDIA A100) and the total training time in Appendix A.4.
    
\item {\bf Code of ethics}
    \item[] Question: Does the research conducted in the paper conform, in every respect, with the NeurIPS Code of Ethics \url{https://neurips.cc/public/EthicsGuidelines}?
    \item[] Answer: \answerYes{}
    \item[] Justification: Our research follows the NeurIPS Code of Ethics.

\item {\bf Broader impacts}
    \item[] Question: Does the paper discuss both potential positive societal impacts and negative societal impacts of the work performed?
    \item[] Answer: \answerNA{}
    \item[] Justification: This work does not explicitly address potential societal impacts.
    
\item {\bf Safeguards}
    \item[] Question: Does the paper describe safeguards that have been put in place for responsible release of data or models that have a high risk for misuse (e.g., pre-trained language models, image generators, or scraped datasets)?
    \item[] Answer: \answerNA{}
    \item[] Justification: Our work does not involve releasing models that pose a high risk of misuse.

\item {\bf Licenses for existing assets}
    \item[] Question: Are the creators or original owners of assets (e.g., code, data, models), used in the paper, properly credited and are the license and terms of use explicitly mentioned and properly respected?
    \item[] Answer: \answerYes{}
    \item[] Justification: The creators or original owners of assets (e.g., code, data, models), used in the paper, are properly credited.


\item {\bf New assets}
    \item[] Question: Are new assets introduced in the paper well documented and is the documentation provided alongside the assets?
    \item[] Answer: \answerNA{}
    \item[] Justification: The paper does not release new assets.

\item {\bf Crowdsourcing and research with human subjects}
    \item[] Question: For crowdsourcing experiments and research with human subjects, does the paper include the full text of instructions given to participants and screenshots, if applicable, as well as details about compensation (if any)? 
    \item[] Answer: \answerNA{}
    \item[] Justification: The paper does not involve crowdsourcing nor research with human subjects.

\item {\bf Institutional review board (IRB) approvals or equivalent for research with human subjects}
    \item[] Question: Does the paper describe potential risks incurred by study participants, whether such risks were disclosed to the subjects, and whether Institutional Review Board (IRB) approvals (or an equivalent approval/review based on the requirements of your country or institution) were obtained?
    \item[] Answer: \answerNA{}
    \item[] Justification: The paper does not involve crowdsourcing nor research with human subjects.

\item {\bf Declaration of LLM usage}
    \item[] Question: Does the paper describe the usage of LLMs if it is an important, original, or non-standard component of the core methods in this research? Note that if the LLM is used only for writing, editing, or formatting purposes and does \emph{not} impact the core methodology, scientific rigor, or originality of the research, declaration is not required.
    %this research? 
    \item[] Answer: \answerNA{}
    \item[] Justification: LLM is used only for writing, editing, or formatting purposes.

\end{enumerate}

\end{document}